\documentclass[12pt,a4paper]{article}

\usepackage[british]{babel}

\usepackage[a4paper,top=2cm,bottom=2cm,left=2.5cm,right=2.5cm,marginparwidth=1.75cm]{geometry}

\usepackage[numbers]{natbib}
\usepackage{multirow}
\usepackage{amsmath}
\usepackage{graphicx}
\usepackage[colorlinks=true, allcolors=blue]{hyperref}
\usepackage{hyperref}
\usepackage{multirow}
\usepackage[title]{appendix}
\usepackage{mathrsfs}
\usepackage{amsfonts}
\usepackage{booktabs} 
\usepackage{caption}  
\usepackage{threeparttable} 
\usepackage{algorithm}
\usepackage{algorithmicx}
\usepackage{algpseudocode}
\usepackage{listings}
\usepackage{enumitem}
\usepackage{chngcntr}
\usepackage{booktabs}
\usepackage{lipsum}
\usepackage{subcaption}
\usepackage{authblk}
\usepackage[T1]{fontenc}    
\usepackage{csquotes}       
\usepackage{diagbox}
\usepackage{subcaption}
\usepackage{graphicx}
\usepackage{float}
\usepackage{caption}
\usepackage{subcaption}
\usepackage{tikz}
\usepackage{amsmath}
\usepackage{color}
\usepackage{gensymb}

\usepackage[symbol]{footmisc}

\usepackage{setspace}
\usepackage{titlesec}
\titleformat{\section} 
  {\normalfont\Large\bfseries}{\thesection.}{1em}{}
  
\usepackage{lineno} 

\rightlinenumbers 

\usepackage{float}   
\usepackage{caption} 


\title{Low-Cost High-Order Singular Value Decomposition for Tensor-Based Reconstruction from Sparse Sensor Measurements: Urban Flow and Air-Quality Applications}

\author[1,*]{Arindam Sengupta}
\author[1,2]{Paul Jeanney}
\author[3]{Ricardo Vinuesa}
\author[1]{Jose Miguel Pérez}
\author[1,*]{Soledad Le Clainche}

\affil[1]{\small ETSI Aeronáutica y del Espacio, Universidad Politécnica de Madrid, Plaza Cardenal Cisneros, 3, Madrid, 28040, Spain}
\affil[2]{\small Ove Arup, C. de Alfonso XI, 12, Retiro, Madrid, 28014, Spain}
\affil[3]{\small Department of Aerospace Engineering, University of Michigan, Ann Arbor, MI 48109, United States}
\affil[*]{Corresponding authors: \texttt{a.sengupta@upm.es} (Arindam Sengupta), \texttt{soledad.leclainche@upm.es} (Soledad Le Clainche) \\

Co-authors:
\texttt{paul.jeanney@arup.com}\\
\texttt{rvinuesa@umich.edu}\\
\texttt{josemiguel.perez@upm.es}}

\date{}

\begin{document}
\maketitle

\begin{abstract} 

Urban flow and air-quality simulations generate high-dimensional datasets describing velocity and pollutant transport across multiple spatial, temporal, and physical-variable dimensions. Reconstructing these fields from sparse sensor measurements is a fundamental challenge in environmental monitoring, digital twins, forecasting, and data assimilation. Existing low-cost reconstruction approaches are commonly based on matrix decompositions, which require multidimensional datasets to be flattened into two-dimensional snapshot matrices, thereby discarding important structural information. This work introduces the low-cost High-Order Singular Value Decomposition (lcHOSVD), a novel tensor-based sparse-sensing reconstruction framework for high-dimensional environmental fields. To the authors’ knowledge, this is the first methodology that combines sparse sensing and HOSVD for field reconstruction. Unlike matrix-based approaches, lcHOSVD preserves the natural tensor structure of the data, enabling the exploitation of correlations across spatial, temporal, and physical-variable dimensions while substantially reducing the computational requirements of conventional HOSVD. The methodology is applied to urban flow and air-quality datasets, where three-dimensional velocity and pollutant concentration fields are reconstructed using only 1–4\% of the available spatial locations. A systematic comparison with low-cost Singular Value Decomposition (lcSVD) is performed to assess the trade-offs between matrix- and tensor-based formulations. While lcSVD provides larger computational speed-ups, lcHOSVD consistently achieves lower reconstruction errors in configurations characterized by strong multidimensional coupling and heterogeneous dynamics across dimensions. Additional sensor-anisotropy analyses demonstrate that the tensor formulation is significantly more robust to uneven sensor distributions, a common situation in practical environmental monitoring networks. The proposed framework extends sparse-sensing reduced-order modelling to tensor representations and demonstrates significant potential for environmental monitoring, forecasting, digital twins, and data-assimilation applications, where complete environmental fields must be estimated from limited observations.

\end{abstract}



\section{Introduction}
\label{sec:intro}

Road traffic is one of the leading sources of NO$_x$ emissions in European cities, releasing nitrogen oxides directly at street level in densely populated areas~\cite{EEA_mobility2024, EEA2024AQ}. Particulate matter (PM$_{2.5}$) from traffic and combustion sources compounds the problem, exposing 96\% of the EU urban population to concentrations above WHO guideline levels~\cite{EEA2024AQ}. How these pollutants accumulate at street level is governed not only by emission rates but also by the surrounding urban geometry and the local wind field. High-fidelity computational fluid dynamics (CFD) can simulate this behaviour, resolving the velocity field, turbulent kinetic energy, and pollutant transport together across domains that range from individual buildings to entire urban districts~\cite{blocken2015computational,tominaga2016ten}. The resulting datasets are no longer simple scalar snapshots, but they are multi-variable, three-dimensional, and, in the time-resolved case, extremely large. Modern fluid or urban flow databases, whether time-resolved particle image velocimetry (PIV) measurements \cite{fang2018particle,mendez2020multiscale}, large-eddy simulations (LES) of reactive flows \cite{bres2018importance,vervisch1998direct}, and high-fidelity CFD simulations \cite{blocken2015computational,clainche2018spatio, tominaga2016ten}, all produce datasets whose sheer size renders direct manipulation computationally expensive. While these databases contain rich physical information about the dominant flow patterns and pollutant pathways, their dimensionality creates bottlenecks in terms of storage, computational cost, and real-time applicability.

Modal decomposition methods have long offered a principled response to these challenges. Proper Orthogonal Decomposition (POD)/ Singular Value Decomposition (SVD)~\cite{golub1971singular, lumley1967structure} remains the most widely used technique for flow analysis, as it provides the low-rank approximation of the data. Its applications span turbulence analysis, flow reconstruction, predictions, and the construction of low-dimensional dynamical models~\cite{berkooz1993proper,holmes2012turbulence, abadia2022predictive}. In urban flow studies, Liu~et~al.~\cite{liu2023proper} applied POD to large-eddy simulation data over real urban morphology and demonstrated that a small number of modes is sufficient to capture the dominant wind patterns, while Xiang~et~al.~\cite{xiang2021non} built a non-intrusive reduced-order model for urban airflow with dynamic boundary conditions by coupling POD with regression techniques, achieving good agreement with full CFD solutions at a fraction of the computational cost. Xiao et al.~\cite{xiao2019reduced} developed a non-intrusive reduced-order model for turbulent urban flows using Gaussian process regression combined with POD, demonstrating predictions several orders of magnitude faster than the full LES solver. Dynamic Mode Decomposition (DMD)~\cite{schmid2010dynamic} is another commonly used modal decomposition method, which extends modal analysis by associating each mode with a specific temporal frequency, making it particularly effective for identifying coherent flow structures in unsteady and turbulent problems. 

Modal decomposition techniques have been widely used to identify dominant flow structures and develop reduced-order predictive models \cite{ding2021reduced, vervecken2015stable, wu2026llm}. Urban flow datasets typically comprise velocity and scalar fields defined over large three-dimensional domains, where the flow behaviour is strongly influenced by building layouts and street-canyon interactions. Modal decomposition methods can capture the principal dynamics of these urban wind fields while significantly reducing the dimensionality of the data \cite{xiao2019reduced, liu2023proper, xiang2021non}. More broadly, Masoumi-Verki et al.~\cite{masoumi2022review} reviewed recent developments in reduced-order modelling for urban airflow and pollutant dispersion, identifying reduced-order models as the practical alternative to CFD simulations and noting that the high computational cost of full CFD simulations prevents their use in near real-time and long-term applications.

Despite their success, conventional POD and DMD approaches typically require the data to be arranged into a two-dimensional snapshot matrix, obtained by flattening spatial dimensions and physical variables into a single vector representation. Although this reshaping preserves the underlying information, it does not explicitly exploit the inherent multiway structure of the data. As a result, spatial directions and physical variables are treated collectively within a single index, preventing the direct identification of independent modal bases associated with each dimension and variable. Tensor-based decompositions provide a direct solution by representing the data as a multiway array and decomposing it along each mode independently. These methods preserve the directional structure of the flow and enable mode bases to be computed separately for each spatial direction and variable. Among them, the High-Order SVD (HOSVD)~\cite{de2000multilinear,de2000best} provides a multilinear generalization of the matrix SVD that extends naturally to tensors. In the Tucker decomposition form~\cite{kolda2009tensor}, HOSVD produces a core tensor and a set of orthonormal factor matrices, one per mode, enabling independent rank selection along each axis. These properties make HOSVD attractive for fluid dynamics datasets, where the dominant complexity may differ significantly between, say, the streamwise and wall-normal directions. Also, Higher-Order Dynamic Mode Decomposition (HODMD) was later introduced by Le Clainche \& Vega~\cite{le2017higher} by generalising the snapshot matrix to include $d$ consecutive delayed time steps, enabling the identification of dynamics that a single-step decomposition cannot resolve.

Although tensor-based decompositions offer a structurally richer representation of multi-dimensional flow data, their practical use is often limited by the high computational cost of full tensor decompositions. Computing HOSVD requires performing a large matrix SVD along each unfolding of the tensor, and as the spatial resolution or number of variables grows, this cost accumulates rapidly across modes. For the kind of high-resolution urban or reactive flow datasets increasingly common in research, even a single unfolding may involve matrices too large and require high-capacity resources.

A complementary strategy to reduce computational overhead comes from using sparse measurements. Instead of decomposing the full dataset, the decomposition can be performed on a reduced subset of the data, and then the full solution can be reconstructed \cite{hetherington2025low}. In lcSVD, SVD is applied to a reduced snapshot matrix formed by selecting a small number of spatial locations, either randomly, equidistantly, or via optimal sensor placement \cite{manohar2018data,de2021pysensors}, and the full-resolution spatial modes and temporal coefficients are then recovered by projecting back onto the original spatial domain. Hetherington \& Le Clainche~\cite{hetherington2025low} introduced and validated the method on a range of test cases spanning laminar and turbulent flows in two and three dimensions, reporting speed-up factors of up to 630 times compared to classical SVD and memory reductions of approximately 37\%, with reconstruction errors comfortably below 5\% for laminar configurations. A different route is taken by Nav et al. \cite{nav2026hierarchical}, who reconstructed sparse-sensor wind fields through a hierarchical, learning-based pipeline. A single sparse-to-fine inversion is replaced by a sequence of coarse-to-fine resolution upgrades. Each learned with an LSTM in a POD coefficient space, with sensors placed by QR pivoting. The two strategies differ mainly in how the unobserved information is restored. lcSVD obtains the full field in one projection step from the sampled modes. In contrast, the hierarchical method restores it progressively through trained surrogates that must be calibrated offline for each resolution transition and scenario. lcSVD relies on a single modal basis. In contrast, the hierarchical approach maintains a separate POD basis and a learned mapping at every level. Recent diffusion-based methods have also explored sparse reconstruction from a different perspective. Diff-SPORT~\cite{vishwasrao2025diff} combines a conditional generative diffusion model for turbulent-flow reconstruction with an explainable-deep-learning strategy for optimal sensor placement, where the sensor locations are identified from the features the network itself deems most informative. This pairing of generative reconstruction with interpretable placement further highlights the growing interest in sparse sensing for urban-flow applications.

Pillai~et~al.~\cite{pillai2025low} applied lcSVD to reactive combustion databases, demonstrating its capacity to reconstruct POD modes from sparse sensors and merge heterogeneous numerical and experimental datasets. For the laminar coflow flame case, reconstruction accuracy remained within 1\% for the dominant species, while the method ran more than ten times faster than standard SVD and reduced the data volume by a factor exceeding 2000. The results presented confirmed that the low-cost framework holds up even under strong multivariate coupling and experimental noise. The lcSVD framework was subsequently extended to data assimilation. Jeanney et al.~\cite{jeanney2025ensemble} applied low-cost Singular Value Decomposition (lcSVD) within a data assimilation framework for fluid dynamics. Their study showed that reduced-resolution computations, combined with lcSVD-based reconstruction, can substantially reduce computational time and memory usage compared to the traditional methods. Just like SVD, lcSVD is effective at reducing computational cost, but it does not preserve the structure of the data. A tensor-based low-cost approach would address this issue by retaining the structural advantages of HOSVD and operating at a reduced cost enabled by sparse sampling.

This paper introduces the low-cost High-Order SVD (lcHOSVD) framework to address this gap. Building on the theoretical foundations of HOSVD and the practical advances demonstrated by lcSVD, the proposed method performs the full decomposition directly from sparse sensor observations, recovering the complete mode bases and core tensor through projection onto the original domain. In this way, the computational cost of conventional HOSVD is substantially reduced while its representation is preserved, with independent mode bases along each spatial direction and physical variable. To the authors' knowledge, this is the first methodology combining sparse sensing and HOSVD for field reconstruction, and the first application of low-cost tensor decomposition techniques to urban flow and pollutant transport datasets. Depending on the test case and on the decay of the singular values, the method can be applied to each variable individually or to the tensor as a whole. A systematic comparison between lcSVD and lcHOSVD is carried out to assess the trade-offs between the matrix- and tensor-based formulations, identifying when the tensor structure provides a meaningful advantage in accuracy and under anisotropic sensor distributions, and extending sparse-sensing reduced-order modelling to tensor representations. The framework is validated on two urban flow configurations of increasing geometric complexity: a single-snapshot, multi-variable dataset of the Vallecas urban district in Madrid, and a time-resolved velocity dataset of the turbulent flow around two wall-mounted buildings.

The remainder of this paper is organized as follows. Section~\ref{sec:methodology} details the lcHOSVD formulation and 
its relationship to HOSVD and lcSVD. Section~\ref{sec:datasets} describes the datasets used for validation. Section~\ref{sec:results} presents reconstruction accuracy, computational performance, and a comparison between methods. The main conclusions are drawn in Section~\ref{sec:conclusions}.

\section{Methodology}
\label{sec:methodology}

\begin{figure}[h]
    \centering
    \includegraphics[width=0.75\textwidth]{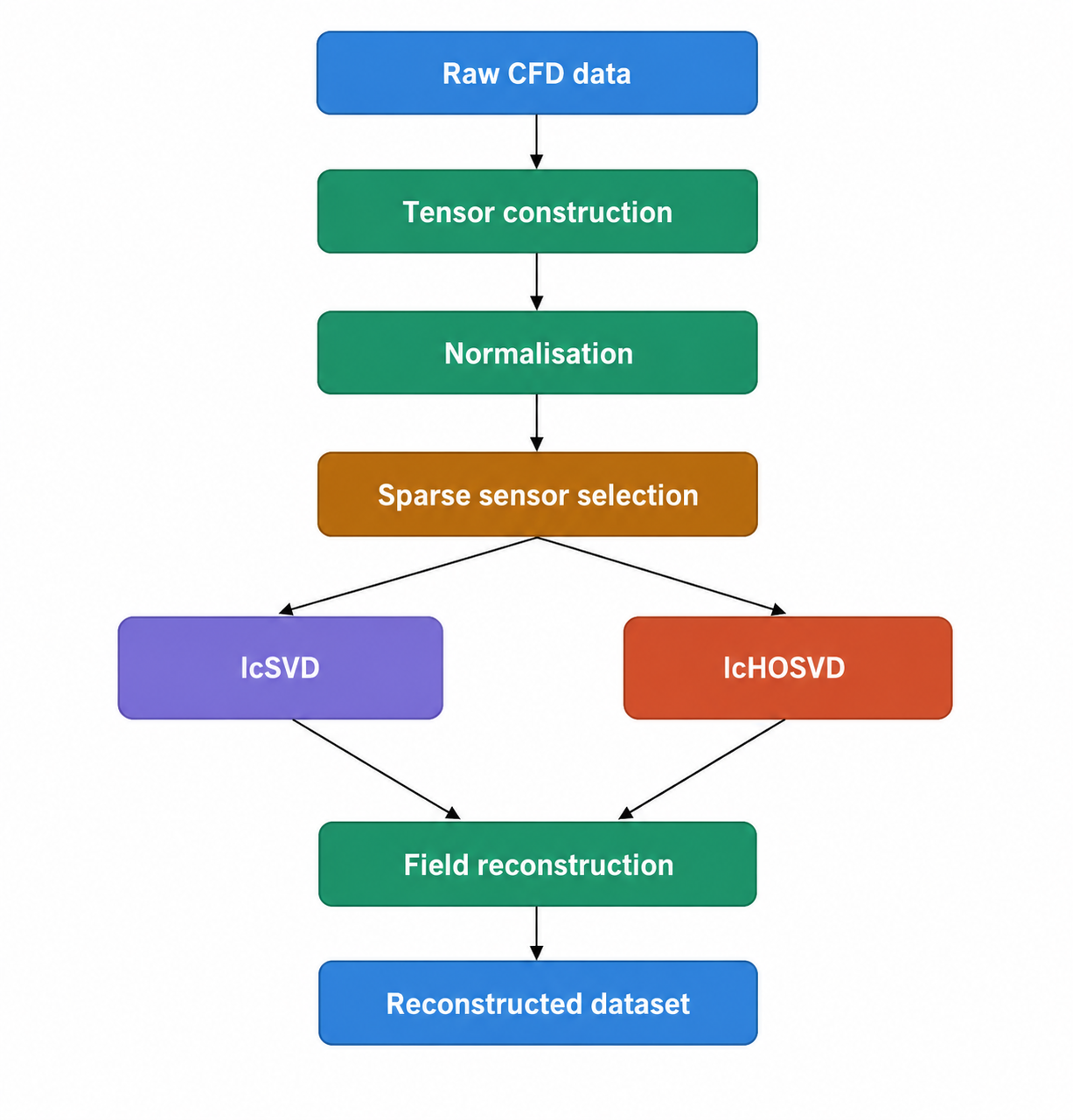}
    \caption{Schematic overview of the methodology: from raw CFD data through tensor construction, normalisation, low-cost 
    decomposition (lcSVD and lcHOSVD), and reconstruction.}
    \label{fig:methodology_schematic}
\end{figure}

This section describes the complete methodological pipeline adopted in the present work. The data are first organised into a structured tensor representation that preserves the multiway nature of the flow fields. Two low-cost decomposition strategies are then applied, the matrix-based lcSVD and the tensor-based lcHOSVD introduced here. Before decomposition, all variables are normalised (Eq.~(\ref{eq:normalization})) to ensure comparability across physical quantities of different magnitudes. The reconstruction quality and flow structure are assessed through the relative root mean square error (RRMSE) and the Q-criterion. A schematic overview of the full pipeline is shown in Fig.~\ref{fig:methodology_schematic}.

\subsection{Data organisation}
\label{sec:data_org}

The datasets considered in this work consist of time-resolved flow fields represented on structured spatial grids. Following the classical snapshot approach, the data can be arranged as a collection of flow states,

\begin{equation}
\boldsymbol{X} = \boldsymbol{V}_1^K =
[\boldsymbol{V}_1,\boldsymbol{V}_2,\ldots,
\boldsymbol{V}_k,\boldsymbol{V}_{k+1},
\ldots,\boldsymbol{V}_K],
\label{eq:snapshot_matrix}
\end{equation}

where $\boldsymbol{V}_k$ denotes the complete flow field at time instant $t_k$, and $K$ is the total number of snapshots. In the conventional matrix formulation, each snapshot is reshaped into a column vector, producing a data matrix of dimensions $J \times K$, where $J$ denotes the total number of spatial degrees of freedom. For three-dimensional flow fields containing multiple physical variables,

\begin{equation}
J = T_{\mathrm{var}}\,\times N_x \times N_y \times N_z,
\end{equation}

where $T_{\mathrm{var}}$ is the number of variables and $N_x$, $N_y$, and $N_z$ are the numbers of grid points in the streamwise, wall-normal, and spanwise directions, respectively.

For three-dimensional problems, the data are represented as a fifth-order tensor \cite{hetherington2024modelflows}:

\begin{equation}
\mathcal{V}
\in
\mathbb{R}^{T_{\mathrm{var}}
\times N_x
\times N_y
\times N_z
\times K},
\label{eq:tensor}
\end{equation}

where the first index corresponds to the physical variables, the next three correspond to the spatial directions, and the final index contains the temporal snapshots.

Each tensor entry is defined as:

\begin{equation}
\mathcal{V}_{i,j_2,j_3,j_4,k}
=
\phi_i
\left(
x_{j_2},
y_{j_3},
z_{j_4},
t_k
\right),
\label{eq:tensor_element}
\end{equation}

where $i \in {1,\ldots,T_{\mathrm{var}}}$ indexes the physical variable $\phi_i$, $j_2$, $j_3$, and $j_4$ denote the spatial locations in the $x$, $y$, and $z$ directions, respectively, and $k$ denotes the temporal snapshot.

For flow datasets with three velocity components along streamwise ($u$), normal ($v$), and spanwise directions ($w$), the tensor can be expressed as:

\begin{equation}
\phi_1=u,
\qquad
\phi_2=v,
\qquad
\phi_3=w,
\end{equation}

yielding a tensor of dimensions

\begin{equation}
\mathcal{V}
\in
\mathbb{R}^{3
\times N_x
\times N_y
\times N_z
\times K}.
\label{eq:last}
\end{equation}

\subsection{Normalisation}
\label{sec:normalisation}

Urban flow databases typically contain variables with substantially different physical scales. For example, velocity components expressed in m/s and pollutant concentrations expressed in kg/m$^3$ may differ by several orders of magnitude. Without normalisation, variables with larger absolute values dominate the decomposition, and physically less energetic but equally important fields are effectively suppressed. To ensure all variables contribute comparably to the modal analysis, each variable $\phi_i$ is normalised independently using min-max scaling~\cite{patro2015normalization}, given by:

\begin{equation}
    \tilde{\phi}_i = \frac{\phi_i - \min(\phi_i)}
    {\max(\phi_i) - \min(\phi_i)},
    \label{eq:normalization}
\end{equation}

where $\phi_i$ is the original field of the $i$-th variable, $\min(\phi_i)$ and $\max(\phi_i)$ are its global minimum and 
maximum over all spatial points and time instants, and $\tilde{\phi}_i \in [0,1]$ is the resulting normalised field. 
The inverse transformation is applied after reconstruction to recover the physical values.

\subsection{Low-cost decomposition techniques}
\label{sec:lcmd}
\begin{figure}[t]
    \centering
    \includegraphics[width=0.75\textwidth]{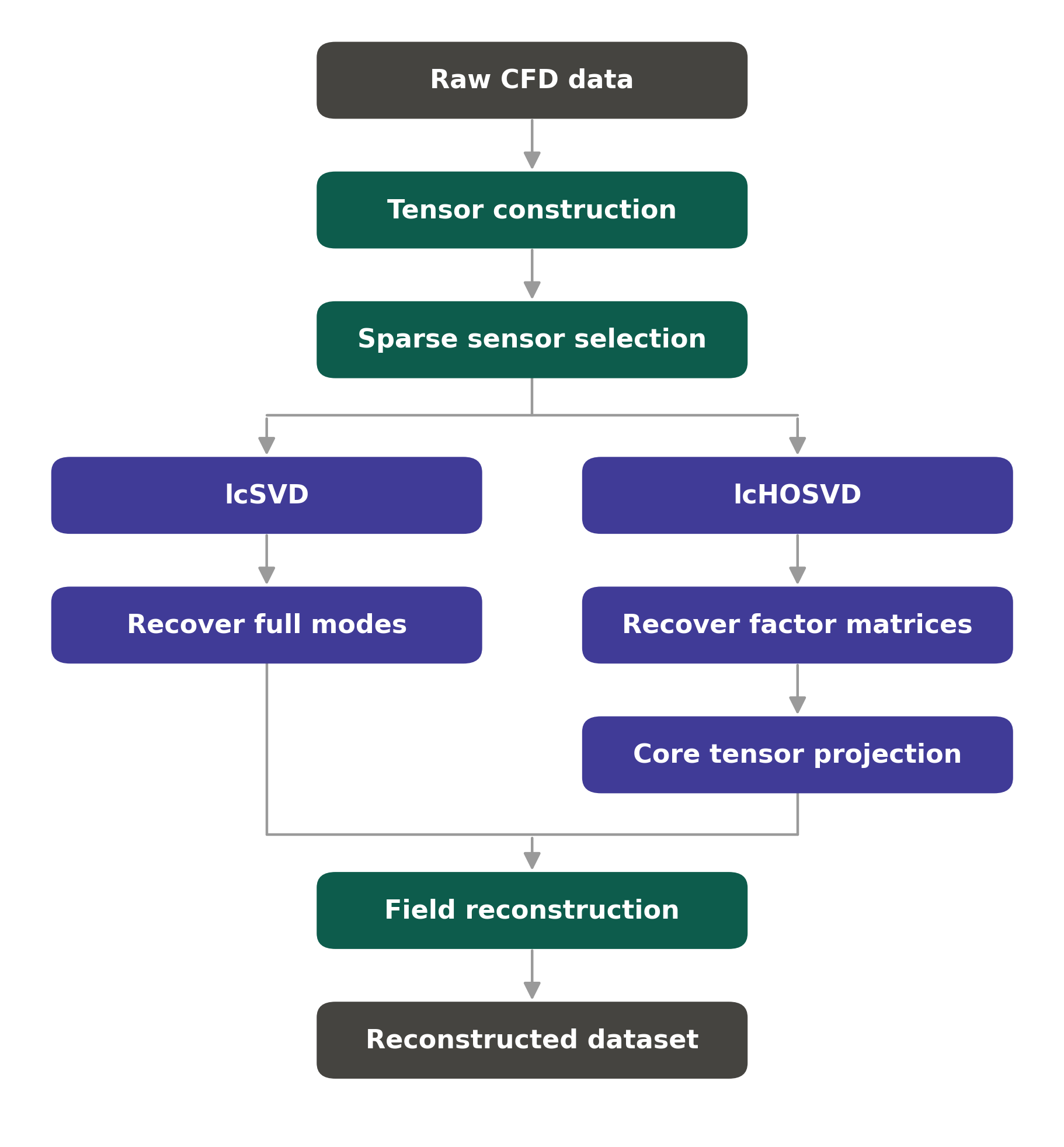}
    \caption{Schematic overview of the lcHOSVD and lcSVD methodology pipelines: from raw CFD data through tensor construction, normalisation, sparse sensor selection, low-cost decomposition, and field reconstruction.}
    \label{fig:methodology1}
\end{figure}

This section presents the two low-cost decomposition strategies compared in this work, the matrix-based lcSVD, introduced by Hetherington \& Le Clainche \cite{hetherington2025low}, and the tensor-based lcHOSVD proposed here. Figure \ref{fig:methodology1} illustrates the low-cost pipeline.

\subsubsection{Low-cost singular value decomposition (lcSVD)}
\label{sec:lcsvd}

In lcSVD, the key idea is to perform the decomposition on a spatially reduced version of the data and subsequently recover the full-resolution spatial modes through projection. Starting from the classical Singular Value Decomposition (SVD),

\begin{equation}
\mathbf{V}_1^K
=\mathbf{U}\,\mathbf{\Sigma}\,\mathbf{T}^{\top},
\label{eq:lcsvd_svd}
\end{equation}

where $\mathbf{U}$ contains the spatial modes, $\mathbf{\Sigma}$ contains the singular values, and $\mathbf{T}^{\top}$ contains the temporal modes. The lcSVD methodology proposed by Hetherington \& Le Clainche~\cite{hetherington2025low} builds on classical SVD (Eq.~(\ref{eq:lcsvd_svd})) by applying the decomposition to a spatially sampled representation of the dataset. The procedure consists of the following steps:

\textbf{1. Construction of the reduced snapshot matrix:}
A subset of $\bar{J}$ spatial locations is selected from the full domain, either equidistantly, randomly, or via optimal sensor placement. 
The reduced snapshot matrix is then:

\begin{equation}
    \bar{\mathbf{V}}_1^K \in \mathbb{R}^{\bar{J} \times K},
    \label{eq:reduced_matrix}
\end{equation}

where $\bar{J} \ll J$ and each column contains the flow field sampled at the selected sensor locations at time $t_m$.

\textbf{2. SVD of the reduced matrix:}
The singular value decomposition is applied to the reduced matrix (Eq.(~\ref{eq:reduced_matrix})):

\begin{equation}
    \bar{\mathbf{V}}_1^K \approx \bar{\mathbf{U}} 
    \bar{\mathbf{\Sigma}} \bar{\mathbf{T}}^\top,
    \label{eq:svd_reduced}
\end{equation}

where $\bar{\mathbf{U}} \in \mathbb{R}^{\bar{J} \times \bar{N}}$ contains the reduced spatial modes in its columns, 
$\bar{\mathbf{\Sigma}} \in \mathbb{R}^{\bar{N} \times \bar{N}}$ is a diagonal matrix of singular values $\sigma_1 \geq \sigma_2 \geq \cdots \geq \sigma_{\bar{N}} > 0$, and $\bar{\mathbf{T}} \in \mathbb{R}^{K \times \bar{N}}$ contains the temporal coefficients. The number of retained modes $\bar{N}$ is determined by the tolerance criterion:

\begin{equation}
    \frac{\sigma_{\bar{N}+1}}{\sigma_1} \leq \varepsilon_{\text{SVD}},
    \label{eq:svd_tolerance}
\end{equation}

where $\varepsilon_{\text{SVD}}$ is a user-defined threshold.

\textbf{3. Recovery of full-resolution modes and temporal coefficients:}
The spatial modes and temporal coefficients of the full dataset are recovered from two semi-reduced matrices: $\mathbf{V}^{J,\bar{K}}\in\mathbb{R}^{J\times\bar{K}}$, containing the full spatial resolution at the retained snapshot columns, and $\mathbf{V}^{\bar{J},K}\in\mathbb{R}^{\bar{J}\times K}$, containing the sensor rows at all snapshots. Equations~(\ref{eq:wrec}--\ref{eq:trec}) define these two recovery steps:
\begin{equation}
    \mathbf{U}^{\mathrm{rec}} = \mathbf{V}^{J,\bar{K}}\,\bar{\mathbf{T}}\,
    \bar{\boldsymbol{\Sigma}}^{-1},
    \label{eq:wrec}
\end{equation}
\begin{equation}
    \mathbf{T}^{\mathrm{rec}} = \left(\mathbf{V}^{\bar{J},K}\right)^{\!\top}
    \bar{\mathbf{U}}\,\bar{\boldsymbol{\Sigma}}^{-1},
    \label{eq:trec}
\end{equation}
where $\mathbf{U}^{\mathrm{rec}}\in\mathbb{R}^{J\times\bar{N}}$ and $\mathbf{T}^{\mathrm{rec}}\in\mathbb{R}^{K\times\bar{N}}$.

\textbf{4. Reconstruction of the full dataset:}
The full snapshot matrix is reconstructed as:

\begin{equation}
    \mathbf{V}_1^{K,\text{rec}} = 
    \mathbf{U}^{\text{rec}} \bar{\mathbf{\Sigma}} 
    \left(\mathbf{T}^{\text{rec}}\right)^\top,
    \label{eq:reconstruction_lcsvd}
\end{equation}

which is then reshaped back into the tensor form (Eq.(~\ref{eq:tensor})) and denormalised to recover physical values.

\subsection{Low-cost higher-order singular value decomposition (lcHOSVD)}
\label{sec:lchosvd}

While lcSVD operates on the flattened snapshot matrix, the lcHOSVD framework preserves the full structure of the data tensor. The method is based on the standard HOSVD, which decomposes the fifth-order tensor into a core tensor multiplied by one orthonormal factor matrix per mode:
\begin{equation}
    \mathcal{V}_{i j_2 j_3 j_4 k} \simeq
    \sum_{p_1=1}^{P_1}\sum_{p_2=1}^{P_2}\sum_{p_3=1}^{P_3}
    \sum_{p_4=1}^{P_4}\sum_{n=1}^{N}
    \mathcal{S}_{p_1 p_2 p_3 p_4 n}\,
    U^{(\mathrm{var})}_{i p_1}\, U^{(1)}_{j_2 p_2}\,
    U^{(2)}_{j_3 p_3}\, U^{(3)}_{j_4 p_4}\, T_{k n},
    \label{eq:hosvd}
\end{equation}

where $\mathcal{S}$ is the core tensor, $U^{(1)}$, $U^{(2)}$, and $U^{(3)}$ are the spatial factor matrices, and $T$ contains the temporal modes.

In the low-cost variant introduced here, the decomposition is computed from sparse sensor measurements along each axis, and the full-resolution factor matrices are then recovered through projection, in direct analogy with the lcSVD procedure.

\textbf{1. Sensor selection along each axis:}
A subset of sensor locations is selected equidistantly and independently along each spatial axis. Let $\bar{N}_x$, $\bar{N}_y$, $\bar{N}_z$ denote the number of retained points along the $x$, $y$, and $z$ directions respectively, with $\bar{N}_x \ll N_x$, $\bar{N}_y \ll N_y$, $\bar{N}_z \ll N_z$, and let $I_x$, $I_y$, $I_z$ be the corresponding sensor index sets,
\begin{equation}
    I_\alpha \subset \{1, \dots, N_\alpha\}, \qquad
    |I_\alpha| = \bar{N}_\alpha, \qquad \alpha \in \{x, y, z\}.
    \label{eq:reduced_tensor}
\end{equation}
These index sets define the rows retained in each mode unfolding in the following step.

\textbf{2. Mode-$n$ unfolding and SVD:}
For each spatial mode $\alpha \in \{x, y, z\}$, the mode-$\alpha$ unfolding of the data is restricted to the sensor indices $I_\alpha$ along that direction, giving $\bar{\mathbf{V}}_{(\alpha)} \in \mathbb{R}^{\bar{N}_\alpha \times M_\alpha}$, where $M_\alpha$ is the product of all remaining dimensions at full resolution. Each column of $\bar{\mathbf{V}}_{(\alpha)}$ is a mode-$\alpha$ fibre evaluated at the sensor indices. For $\alpha = x$, it is obtained by fixing the indices along $y$, $z$, the variable mode, and the temporal mode, and retaining the values at the $\bar{N}_x$ sensor positions along $x$. SVD is then applied independently to each unfolding:
\begin{equation}
    \bar{\mathbf{V}}_{(\alpha)} \approx
    \bar{\mathbf{U}}_\alpha \bar{\mathbf{\Sigma}}_\alpha
    \bar{\mathbf{T}}_\alpha^\top,
    \label{eq:mode_svd}
\end{equation}
where $\bar{\mathbf{U}}_\alpha \in \mathbb{R}^{\bar{N}_\alpha \times r_\alpha}$ contains the left singular vectors (reduced factor matrix for mode $\alpha$), $\bar{\mathbf{\Sigma}}_\alpha \in \mathbb{R}^{r_\alpha \times r_\alpha}$ is the diagonal matrix of singular values, and $\bar{\mathbf{T}}_\alpha \in \mathbb{R}^{M_\alpha \times r_\alpha}$ contains the right singular vectors. Here $\bar{\mathbf{T}}_\alpha$ spans the remaining dimensions $M_\alpha$ at full resolution and is shared between the sensor-restricted and full-resolution mode-$\alpha$ unfoldings, which is the basis exploited in the lift of Eq.~(\ref{eq:factor_rec}). The rank $r_\alpha$ along each mode is selected independently using the tolerance criterion (Eq.~(\ref{eq:svd_tolerance})), or prescribed directly by the user based on singular value analysis.

\textbf{3. Recovery of full-resolution factor matrices:}
The full-resolution factor matrix for each spatial mode $\alpha$ is recovered by projecting the full-resolution mode-$\alpha$ unfolding $\mathbf{V}_{(\alpha)} \in \mathbb{R}^{N_\alpha \times M_\alpha}$ of the original tensor onto the reduced right singular basis $\bar{\mathbf{T}}_\alpha$ obtained from the reduced unfolding~(Eq. (\ref{eq:mode_svd})):
\begin{equation}
    \mathbf{U}_\alpha^{\text{rec}} = 
    \mathbf{V}_{(\alpha)} \bar{\mathbf{T}}_\alpha 
    \left(\bar{\mathbf{\Sigma}}_\alpha\right)^{-1},
    \label{eq:factor_rec}
\end{equation}
where $\mathbf{U}_\alpha^{\text{rec}} \in \mathbb{R}^{N_\alpha \times r_\alpha}$ is the reconstructed full-resolution factor matrix for mode $\alpha$. This step lifts the reduced factor matrix, computed from the sparse sensor measurements, back to the full spatial resolution $N_\alpha$ by exploiting the right singular vectors as a shared basis between the reduced and full unfoldings.

\textbf{4. Core tensor computation:}
The core tensor $\mathcal{G} \in \mathbb{R}^{n_{\text{var}} \times r_x \times r_y \times r_z \times r_t}$ is obtained by projecting the original tensor $\mathcal{V}$ onto all five reconstructed factor matrices:
\begin{equation}
    \mathcal{G}_{p_1 p_2 p_3 p_4 p_5} = \sum_{i,j,k,l,m} 
    \mathcal{V}_{ijklm} \,
    (U_{\text{var}}^{\text{rec}})_{ip_1} \,
    (U_x^{\text{rec}})_{jp_2} \,
    (U_y^{\text{rec}})_{kp_3} \,
    (U_z^{\text{rec}})_{lp_4} \,
    (U_t^{\text{rec}})_{mp_5},
    \label{eq:core_tensor}
\end{equation}
where $i$, $j$, $k$, $l$, $m$ are the indices along the variable, $x$, $y$, $z$, and temporal modes of $\mathcal{V}$, and $p_1, \ldots, p_5$ are the corresponding compressed indices of $\mathcal{G}$. The factor matrices along the variable and temporal modes, $\mathbf{U}_{\mathrm{var}}^{\mathrm{rec}}$ and $\mathbf{U}_t^{\mathrm{rec}}$, are computed directly from the standard SVD of the corresponding full unfoldings.

\textbf{5. Reconstruction of the full tensor:}
The full-resolution tensor is recovered by expanding the core tensor back through all factor matrices:
\begin{equation}
    \mathcal{V}^{\text{rec}}_{ijklm} = \sum_{p_1,p_2,p_3,p_4,p_5} 
    \mathcal{G}_{p_1 p_2 p_3 p_4 p_5} \,
    (U_{\text{var}}^{\text{rec}})_{ip_1} \,
    (U_x^{\text{rec}})_{jp_2} \,
    (U_y^{\text{rec}})_{kp_3} \,
    (U_z^{\text{rec}})_{lp_4} \,
    (U_t^{\text{rec}})_{mp_5},
    \label{eq:reconstruction_lchosvd}
\end{equation}
where the indices $i$, $j$, $k$, $l$, $m$ now run over the full spatial and temporal dimensions $n_{\text{var}}$, $N_x$, $N_y$, $N_z$, and $K$ respectively, recovering the complete flow field at full resolution.

\begin{figure}[t]
    \centering
    \includegraphics[width=\textwidth]{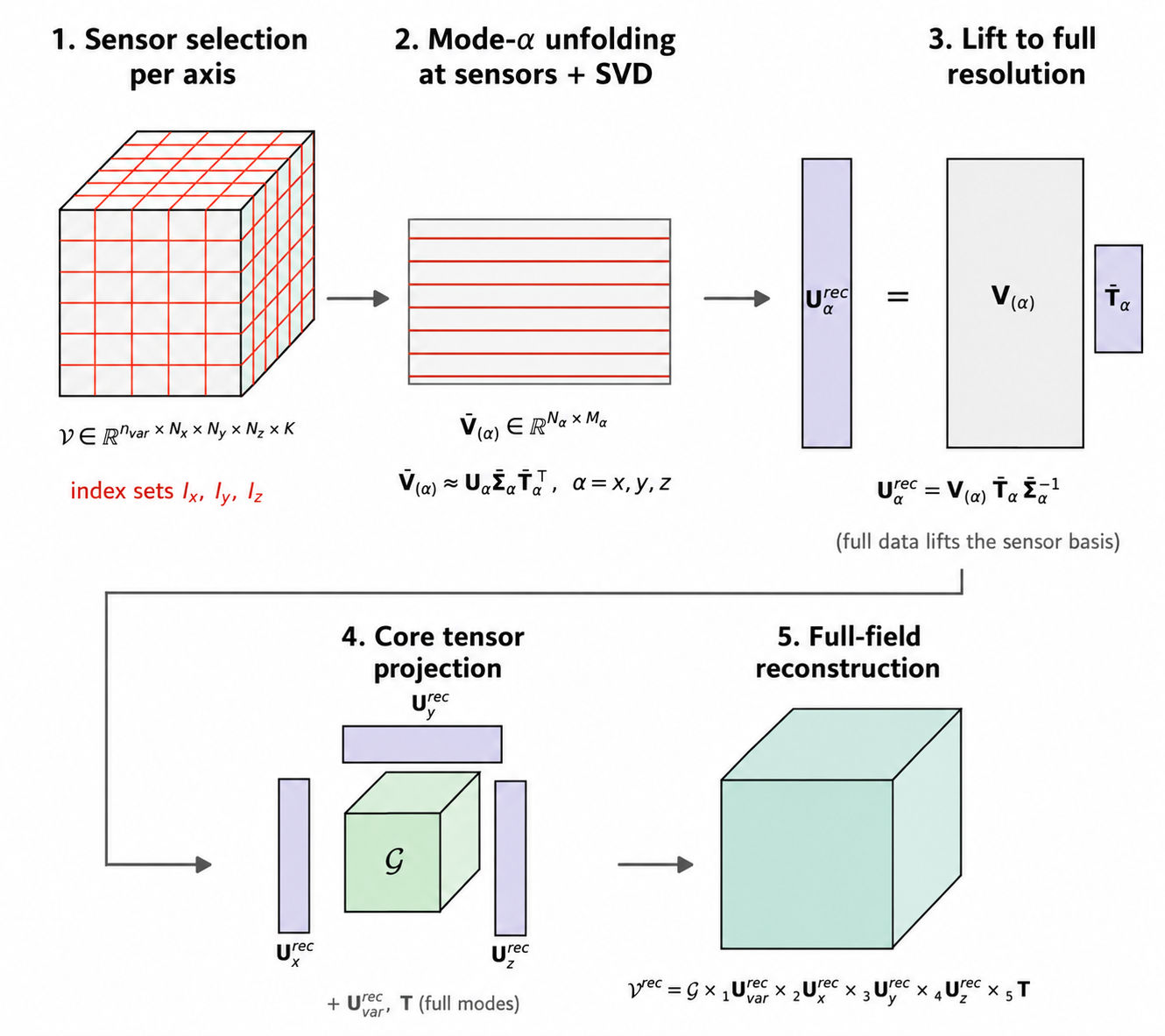}
    \caption{Schematic overview of the steps in the lcHOSVD methodology.}
    \label{fig:method1}
\end{figure}

Step-by-step illustration of the lcHOSVD methodology is presented in Fig.\ref{fig:method1}. Sensors are selected independently along each spatial axis (red planes). Each mode-$\alpha$ unfolding is restricted to the sensor rows and decomposed, and the full-resolution factor matrices are recovered by lifting through the right singular basis. The core tensor is obtained by projection onto all factor matrices, and the full field is reconstructed by expanding the core through the factor bases. 

This formulation also admits a direct extension to data assimilation, since the combination of sensor information and CFD data is already built into Eqs.~(\ref{eq:factor_rec}) and~(\ref{eq:core_tensor}). The sensor measurements enter the method through the restricted unfoldings (Eq.~(\ref{eq:mode_svd})), which provide the reduced bases $\bar{\mathbf{T}}_\alpha$ and $\bar{\boldsymbol{\Sigma}}_\alpha$, while the CFD data is processed through the lift (Eq.~(\ref{eq:factor_rec})) and the core projection (Eq.~(\ref{eq:core_tensor})), where the full-resolution fields are combined with the sensor-derived bases. In the present work, the sensor values are extracted from the CFD solution itself, but the same equations apply when the restricted unfoldings are assembled from real observations in place of, or merged with, the CFD values at the sensor locations, following a hybrid-matrix strategy. The lift then propagates the assimilated information to the full spatial resolution through the factor matrix of each direction. Since each spatial mode holds its own restricted unfolding, this extension preserves the tensor structure throughout the assimilation cycle and allows direction-dependent weighting of the observational and model contributions, a possibility that the flattened formulation does not offer.

\subsection{Performance assessment}
\label{sec:metrics}

Three metrics are used to assess reconstruction quality and flow structure: the relative root mean square error (RRMSE) for quantitative accuracy, the Q-criterion for physical assessment of the reconstructed structures, and the compression ratio (CR) for savings.

\subsubsection{Relative root mean square error}

The RRMSE measures the pointwise agreement between the reconstructed field and the reference CFD solution. For a given variable, it is defined as:

\begin{equation}
    \mathrm{RRMSE} =
    \frac{\sqrt{\dfrac{1}{N}\displaystyle\sum_{i=1}^{N}
    \left(\mathbf{V}_i-\mathbf{V}^{\mathrm{rec}}_i\right)^{2}}}
    {\sqrt{\dfrac{1}{N}\displaystyle\sum_{i=1}^{N}\mathbf{V}_i^{2}}},
    \label{eq:rrmse}
\end{equation}
where \(\mathbf{V}_i\) represents the reference values, \(\mathbf{V}_i^{rec}\) the reconstructed values obtained from the low-cost method, and \(\boldsymbol{N}\) the total number of samples. The RRMSE is evaluated exclusively over fluid (non-building) cells, since building cells carry no physical flow information and their inclusion would artificially affect the error metric.

\subsubsection{Q-criterion}

The Q-criterion~\cite{hunt1988eddies} is a widely used approach for identifying vortical structures in reconstructed three-dimensional velocity fields. Although global error metrics provide an overall measure of reconstruction accuracy, they do not necessarily reflect the ability of the reduced-order models to preserve physically relevant flow structures. Therefore, a qualitative comparison based on the Q-criterion is performed to assess the reconstruction. It is defined as the second invariant of the velocity gradient tensor $\nabla \mathbf{u}$:

\begin{equation}
    Q = \frac{1}{2}\left( 
    \|\mathbf{\Omega}\|_F^2 - \|\mathbf{S}\|_F^2 
    \right),
    \label{eq:qcriterion}
\end{equation}

where $\mathbf{\Omega}$ is the antisymmetric rotation rate tensor, $\mathbf{S}$ is the symmetric strain rate tensor, and $\|\cdot\|_F$ denotes the Frobenius norm. Regions where $Q > 0$ indicate that rotation dominates over strain.

\subsubsection{Compression factor}

The compression factor (CF) is defined as the fraction of spatial locations retained relative to the full spatial domain:

\begin{equation}
CF = \frac{N_x N_y N_z}{N_{\mathrm{sensors}}},
\label{eq:sensor_ratio}
\end{equation}

where \(N_{\mathrm{sensors}}\) denotes the number of retained sensor locations and \(N_xN_yN_z\) represents the total number of spatial grid points in the full domain. The corresponding compression percentage is defined as:

\begin{equation}
\mathcal{C(\%)}=\left(1-\frac{N_{\mathrm{sensors}}}{N_xN_yN_z}\right)\times 100\%,
\label{eq:compression}
\end{equation}

where \(\mathcal{C}\) represents the percentage reduction in spatial degrees of freedom.

\section{Datasets}
\label{sec:datasets}

Two urban flow datasets of different flow natures and geometric complexities are considered in this study. A large-scale three-dimensional urban CFD simulation of the Vallecas district in Madrid and a turbulent two-building case computed with high-fidelity LES were used to evaluate the low-cost framework.

\subsection{Three-dimensional urban CFD dataset (Vallecas, Madrid)}
\label{sec:data_vallecas}

The first dataset corresponds to a large-scale urban CFD case study in the Vallecas district of southeastern Madrid \cite{jeanney2026large}, selected because of its dense residential and educational land use and its proximity to the A-3 and M-40 motorway corridors, two of the busiest roads in the Madrid metropolitan area. It is further motivated by the ongoing municipal redevelopment plan that includes the construction of approximately 1,400 housing units and a student residence in the area \citep{ElEconomista2024Vallecas}.

The three-dimensional urban geometry was reconstructed at Level of Detail 2.2 using the open-source tool city4CFD \citep{city4cfd}, which fits the true roof shape of each building from the LiDAR point cloud obtained from the Madrid Geoportal \citep{MadridAQ}. Terrain, buildings, vegetation, and water bodies are represented as distinct surface layers. The computational domain follows the best-practice guidelines of Blocken et al. \citep{blocken2015computational}, with a horizontal extent of approximately 15 times the maximum building height and a vertical extent of 6 times the maximum building height. This yields a domain diameter of approximately 3410\,m and a vertical extent ranging between 317\,m and 358\,m, due to variations in terrain elevation across the domain. The resulting mesh comprises approximately 100 million cells.

Steady-state incompressible RANS simulations (Eq.~\ref{eq:cont} to \ref{eq:scalar}) were carried out with OpenFOAM \citep{openfoam} using the SimpleFoam solver \citep{simplefoam}, which solves the continuity and momentum equations:
\begin{equation}
    \nabla \cdot \mathbf{U} = 0,
    \label{eq:cont}
\end{equation}
\begin{equation}
    \nabla \cdot \left(\mathbf{U}\mathbf{U}\right) =
    -\nabla p + \nabla \cdot
    \bigl[(\nu+\nu_t)(\nabla\mathbf{U}+\nabla\mathbf{U}^T)\bigr].
    \label{eq:mom}
\end{equation}

where $\mathbf{U}$ is the Reynolds-averaged velocity vector, $p$ is the kinematic pressure, $\nu$ is the kinematic viscosity, and $\nu_t$ 
is the turbulent eddy viscosity.

Turbulence closure is provided by the standard $k$--$\varepsilon$ model \citep{launder}, with constants adjusted for a neutral atmospheric boundary layer (ABL) \citep{hargreaves2007use} and custom wall functions \citep{parente}. Inlet profiles follow the logarithmic ABL law \citep{parente}:
\begin{equation}
    U(z)=\frac{u_*}{\kappa}\ln\!\frac{z+z_0}{z_0},\quad
    k=\frac{u_*^2}{\sqrt{C_\mu}},\quad
    \varepsilon=\frac{u_*^3}{\kappa(z+z_0)},
    \label{eq:abl}
\end{equation}
where $u_*$ is the friction velocity derived from the hourly reference wind speed $U_\mathrm{ref}$ measured at $z_\mathrm{ref}=5.5$\,m, and $z_0=0.25$\,m is the aerodynamic roughness length representative of the dense urban fabric.
Distinct values are assigned to vegetated areas ($z_{0,\mathrm{veg}}=0.10$\,m) and water surfaces ($z_{0,\mathrm{water}}=0.002$\,m). A zero-gradient condition is imposed at the outlet and a symmetry condition at the top boundary. The selected domain height is sufficient to minimize the influence of the upper boundary condition on the flow field.

Concentrations of carbon monoxide (CO), nitrogen oxides (NO$_x$), and particulate matter (PM) are modelled by a passive scalar transport equation coupled to the converged velocity field:
\begin{equation}
    \nabla\cdot(\mathbf{U}\phi) =
    \nabla\cdot\!\left[\left(\frac{\nu}{Sc}+\frac{\nu_t}{Sc_t}
    \right)\nabla\phi\right] + S_\phi,
    \label{eq:scalar}
\end{equation}
where $Sc$ and $Sc_t$ are the molecular and turbulent Schmidt numbers, and $S_\phi$ is a volumetric source term derived from
hourly traffic counts on the A-3 and M-40 corridors, disaggregated by vehicle category (passenger cars, light-duty vehicles,
heavy-duty vehicles, motorcycles) from camera-detection data (Dirección General de Tráfico, DGT). Emission factors follow the
COPERT methodology \citep{EMEPEEA2023} and are distributed as line-source intensities over the road-surface cells of the mesh.

The dataset comprises hourly steady-state simulations for the 24 hours
of 1st February 2024. This was selected because it recorded the highest daily pollutant concentrations of the year at the nearest station of the Madrid air-quality monitoring network. Wind conditions on that day are predominantly from the north and northeast, with a mean reference speed of 0.7\,m/s and a maximum of 1.2\,m/s, yielding a Reynolds number of approximately $10^7$. The methodology has been tested on the first hour (00:00 to 01:00 hrs) of data from the hourly steady-state simulations.

\begin{figure}[htbp]
    \centering
    \includegraphics[width=\textwidth]{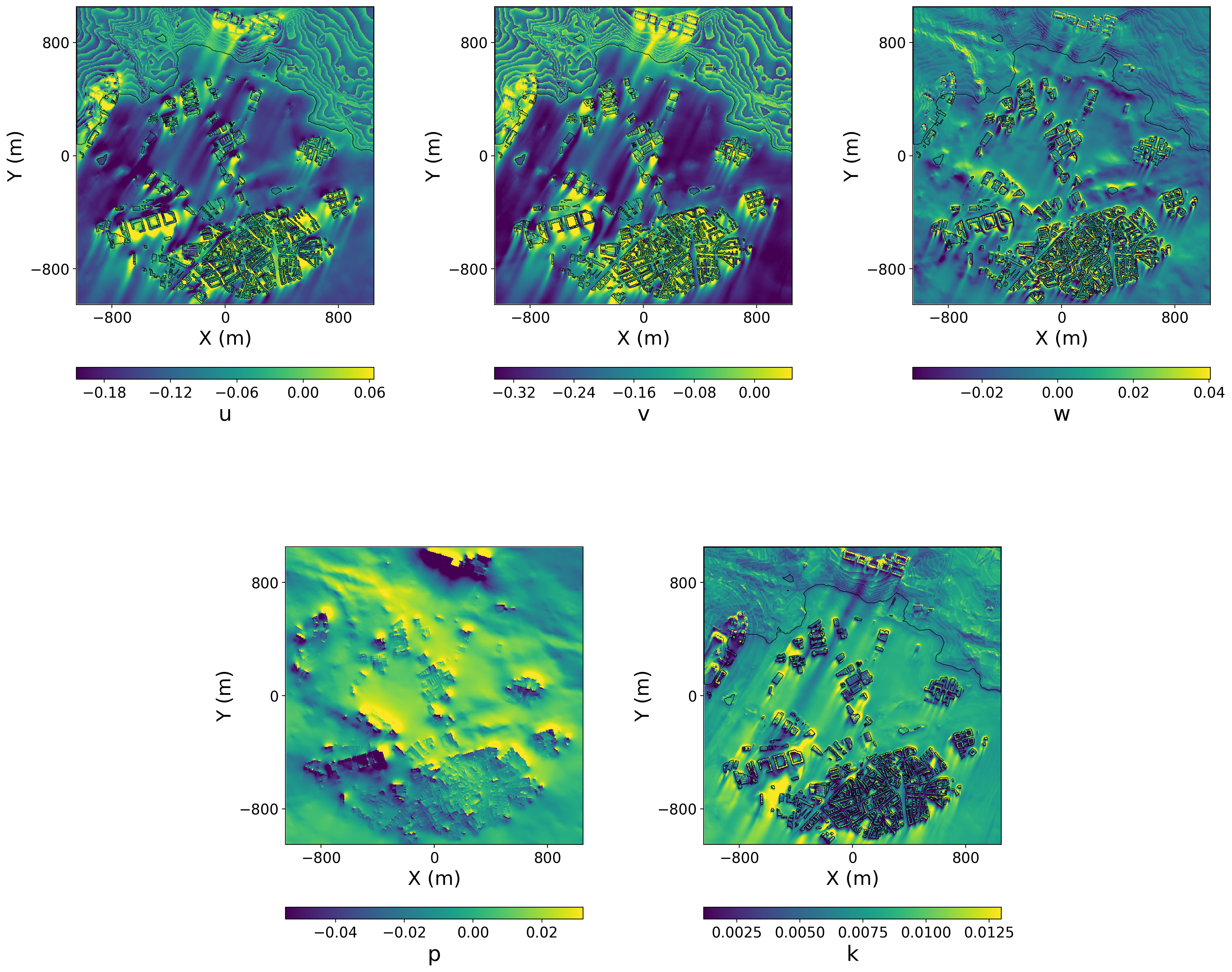}

    \caption{
    Ground-truth velocity fields extracted at AGL (\(z=5\) m) for the Vallecas urban-flow dataset. From left to right (1st row): streamwise, wall-normal, and spanwise velocity components (\(u\) (m/s), \(v\) (m/s), and \(w\) (m/s)). From left to right (2nd row): Kinematic pressure and turbulent kinetic energy (\(p\) (m$^2$/s$^2$) and \(k\) (m$^2$/s$^2$)). The lines mark where terrain or buildings rise above the $z=5$~m AGL plane.}
    
    \label{fig:vallecas_physical_gt_hosvd}
\end{figure}
\begin{figure}[htbp]
    \centering
    \includegraphics[width=0.9\textwidth]{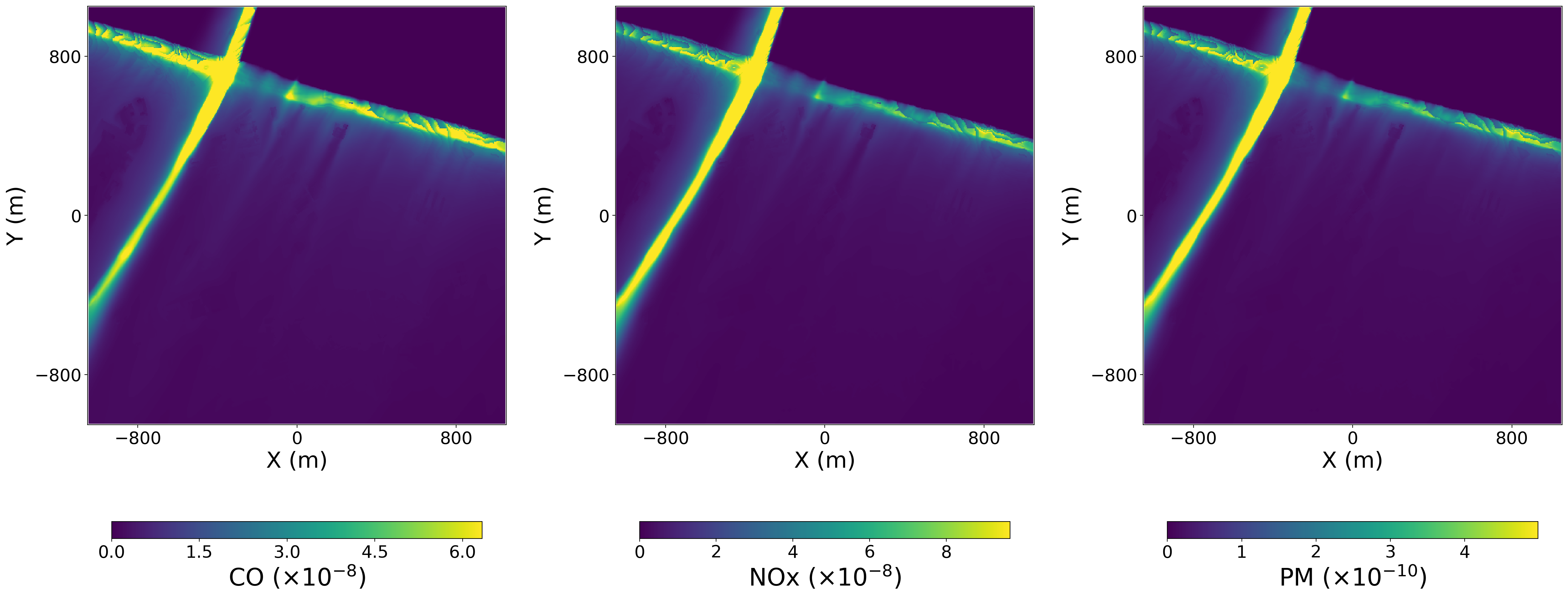}

    \caption{
    Ground-truth pollutant concentration fields extracted at AGL (\(z=5\) m) for the Vallecas urban-flow dataset. From left to right: carbon monoxide, nitrogen oxides, and particulate matter (CO (kg/m$^3$), NO$_x$ (kg/m$^3$), and PM (kg/m$^3$)). 
    }
    \label{fig:vallecas_pollutants_gt_hosvd}
\end{figure}
For the ROM analysis, the data are transferred to a structured Cartesian grid via a terrain-following interpolation. For each horizontal cell $(x_i, y_j)$, the terrain elevation $z_\mathrm{terrain}(x_i,y_j)$ is extracted using a KD-tree nearest-neighbour query \cite{virtanen2020scipy} on the unstructured mesh, and the field variables are then sampled at $N_z = 30$ height levels above ground level (AGL) from 5\,m to 70\,m. This produces a five-dimensional tensor:
\begin{equation}
    \mathbf{V} \in \mathbb{R}^{n_\mathrm{var} \times N_x \times N_y
    \times N_z \times N_t},
    \label{eq:tensor5d}
\end{equation}
where $n_\mathrm{var} = 8$ corresponds to the variables $[u,\,v,\,w,\,p,\,\mathrm{CO},\,\mathrm{NO}_x,\,\mathrm{PM},\,k]$,
$(N_x, N_y) = (500, 500)$ are the horizontal grid dimensions, $N_z = 30$ the vertical levels, and $N_t$ the number of simulation
hours. Here $p$ denotes the kinematic pressure, $k$ the turbulent kinetic energy, and $\mathrm{CO}$, $\mathrm{NO}_x$, $\mathrm{PM}$ the concentrations of carbon monoxide, nitrogen oxides, and particulate matter, respectively. The ground truth snapshots have been presented in Fig.~\ref{fig:vallecas_physical_gt_hosvd} and \ref{fig:vallecas_pollutants_gt_hosvd}, and the pollutant concentrations are displayed scaled by the factor indicated in each colour-bar label ($10^{-8}$~kg/m$^3$ for CO and NO$_x$, and $10^{-10}$~kg/m$^3$ for PM). The lines indicate the boundaries of elevated terrain and building surfaces intersecting the $z = 5$~m horizontal plane, delineating regions where the ground or urban canopy rises above the selected height level. In contrast to data-driven surrogates that must be trained on input-output pairs, the present framework is purely reconstructive. The sparse input is obtained by sampling the reference field at the sensor locations, so for every case considered in this study both the input and the full-resolution reference are known, allowing the reconstruction error to be evaluated directly.

For datasets containing a single temporal snapshot, the rank along the temporal mode is one, which results in an exact reconstruction and provides no information about the dominant spatial structures. To enable a meaningful modal analysis, the spatial dimensions are reshaped into a two-dimensional snapshot matrix. Among the possible arrangements, the $(X, Z) \times Y$ configuration is adopted here, where the streamwise and spanwise directions are combined into a single spatial coordinate of size $N_x N_z$, yielding a matrix of dimensions $N_x N_z \times N_y$.  The $(X, Y) \times Z$ arrangement was not adopted as it places the spanwise direction in the snapshot role, limiting the maximum number of extractable modes to $N_z = 30$. Given the spatial complexity of the horizontal flow field across a $500 \times 500$ grid, this rank ceiling is too restrictive to capture the dominant structures adequately. The arrangement $(Y, Z) \times X$ was also tested and produced comparable results.

\subsection{Two-building urban flow dataset}
\label{sec:data_twobuilding}

The second dataset consists of a DNS simulation of flow around two wall-mounted rectangular obstacles, representing a simplified urban street configuration, which was obtained from the work of Á. Martínez-Sánchez et al. \citep{martinez2023data}. The governing equations and modelling approach differ from those described previously and are detailed in Ref. \citep{martinez2023data}. The simulations were performed using the spectral-element code Nek5000 at a Reynolds number $Re_h = 10{,}000$ based on the obstacle height $h$ and the free-stream velocity $U_\infty$.

The domain consists of two identical wall-mounted obstacles. Each obstacle has height $h$, taken as the reference length and used to define the Reynolds number $Re_h$; all lengths are normalised by $h$. The streamwise length of each obstacle is $w_b$, and its spanwise width is $b$, with $w_b/h = b/h = 0.5$. The obstacles sit in a channel of wall-normal extent $L_y = 3h$ and spanwise extent $L_z = 4h$, and are separated in the streamwise direction by a distance $\ell$. Three separation ratios $\ell/h = 1, 2, 4$ were simulated, corresponding to the skimming-flow (SF), wake-interference (WI), and isolated-roughness (IR) regimes. The tensor used in this study was obtained for the SF case ($\ell/h = 1$).

The mesh employs an eight-point Gauss-Lobatto Legendre (GLL) quadrature within each spectral element, refined near the obstacle surfaces. The resolution satisfies the criteria with the wall-normal and spanwise directions held fixed across all three cases at approximately $6\times 10^6$ grid points for the SF regime. The inflow is located at $x/h = -10$, with the tripping force applied at $x/h = -9$, allowing the boundary layer to develop over the region $-8 \leq x/h \leq -1$, reaching fully turbulent conditions ($Re_\tau \approx 175$, $Re_\theta \approx 450$) upstream of the obstacle at $x/h = -2$.
\begin{figure}[h!]
    \centering

    \includegraphics[width=0.6\textwidth]{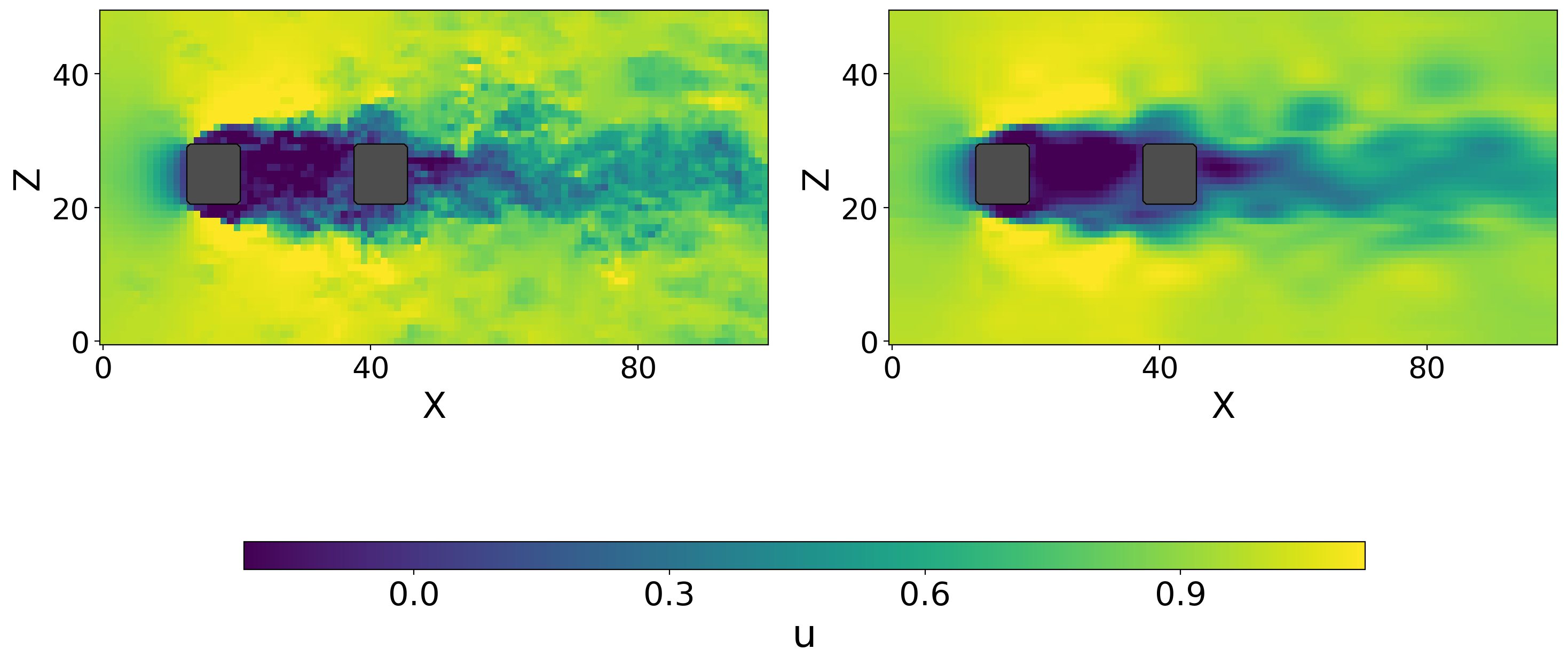}
    \vspace{0.3cm}

    \includegraphics[width=0.6\textwidth]{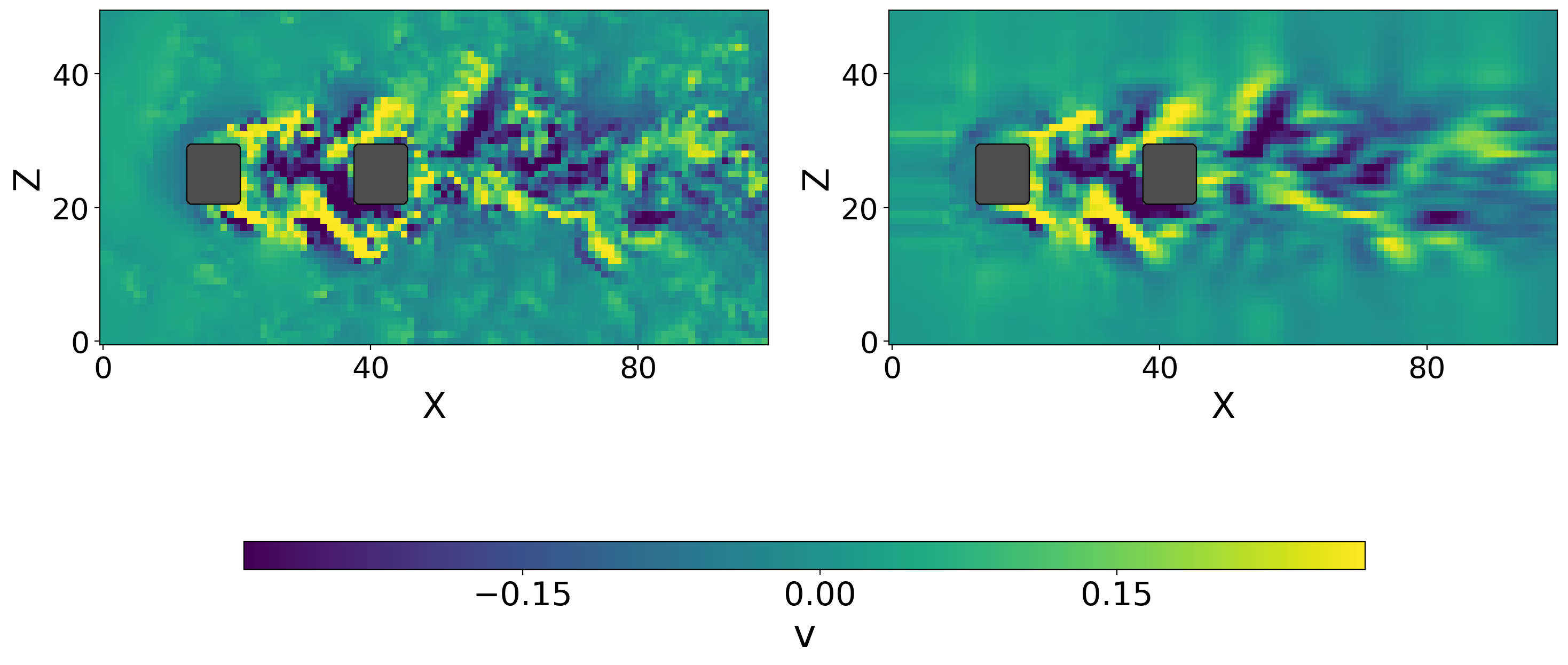}
    \vspace{0.3cm}

    \includegraphics[width=0.6\textwidth]{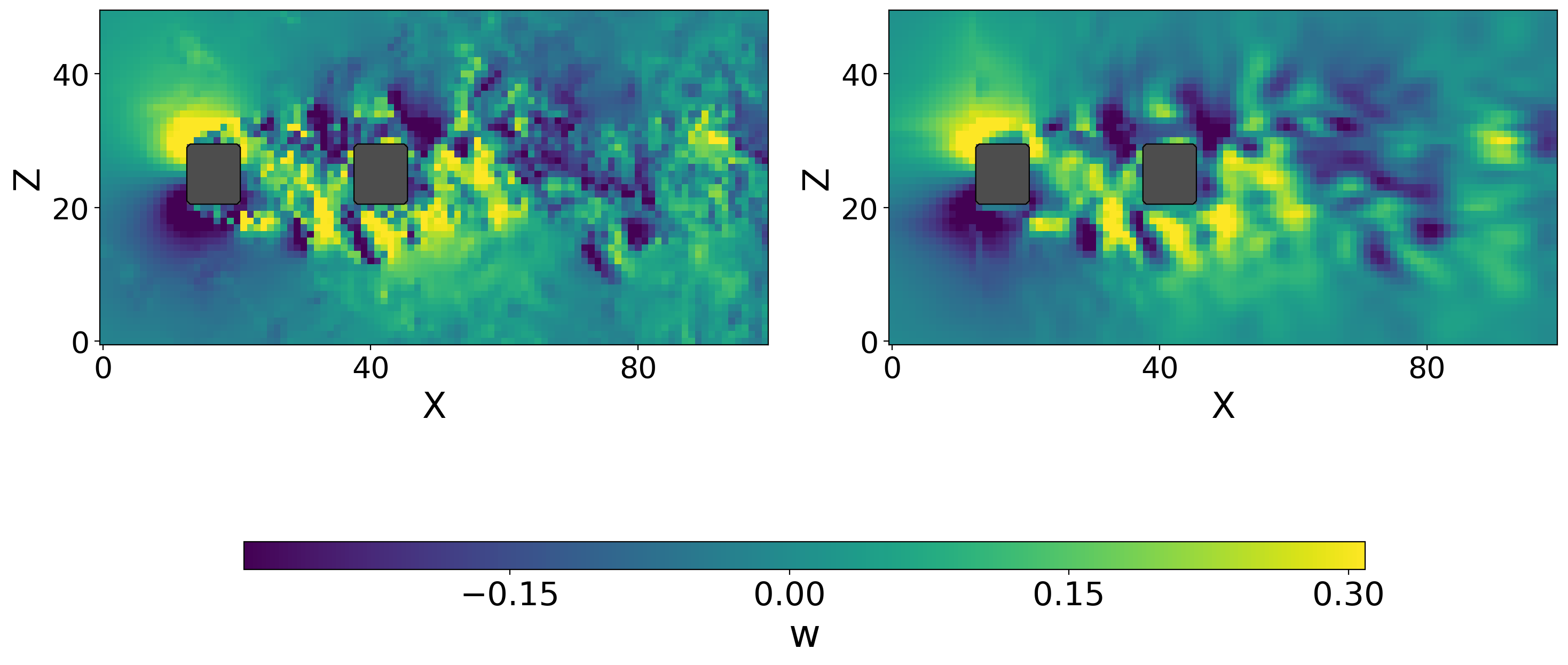}

    \caption{
     Ground-truth (left) and HOSVD-reconstructed (right) snapshots for the two-building  dataset with retained modes $(19,14,15)$. From top to bottom: streamwise velocity $u/U_\infty$, wall-normal velocity $v/U_\infty$, and spanwise velocity $w/U_\infty$. All quantities are non-dimensional: velocities are scaled by the free-stream velocity $U_\infty$, and the spatial axes, shown as $x$ and $y$ in the panels, denote the coordinates normalized by the obstacle height, i.e.\ $x/h$ and $y/h$.}
    
    \label{fig:twobldg_gt_hosvd_snapshots}
\end{figure}

Figure~\ref{fig:twobldg_gt_hosvd_snapshots} shows representative ground-truth and HOSVD-reconstructed snapshots for the three velocity components from the SF regime. The comparison is shown for $u$, $v$, and $w$, providing a direct visual assessment of the reconstruction quality across the different flow components. The raw data contain a broad range of turbulent scales and small-scale fluctuations that are not retained within a reduced-order representation. Since the objective of the present work is to assess the ability of the proposed low-cost methodology to reproduce the dominant flow structures, comparisons for this test case are performed against the HOSVD reconstruction. This provides a consistent basis for evaluation by isolating the effects of the sparse sampling from the discrepancies associated with unresolved small-scale turbulence. The HOSVD reconstruction represents the best approximation attainable at the chosen modes when the full data are available, so it constitutes the natural upper bound for any low-cost variant operating at the same modes. The difference between a low-cost reconstruction and this reference measures only the error introduced by computing the decomposition from the sensor subset. The HOSVD reference is computed with $19, 14, 15$ modes along the $x$, $y$, and $z$ directions, selected from the singular-value decay as discussed in Section~\ref{sec:results_modes}.

Table~\ref{tab:datasets} summarises the two datasets used in this study, including the tensor shape, the physical variables retained, and the number of temporal snapshots.  This generalised formulation accommodates datasets of varying composition. In the two-building flow case considered here, $T_{\mathrm{var}} = 3$ corresponds to the three velocity components, while for the Vallecas urban CFD case, the tensor includes velocity components, pressure, turbulent kinetic energy, and multiple pollutant concentrations, giving a larger $T_{\mathrm{var}}$. In both cases, the structure (Eqs. (\ref{eq:snapshot_matrix}-\ref{eq:last})) remains unchanged; only the number of variable slices in the first tensor mode varies. The tensor shape follows the convention $(n_\text{var} \times N_x \times N_y \times N_z \times N_t)$, where $n_\text{var}$ is the number of physical variables, $N_x$ and $N_y$ are the number of grid points in the streamwise and wall-normal directions, $N_z$ in the spanwise direction, and $N_t$ is the number of temporal snapshots.

\begin{table}[h!]
\centering
\begin{tabular}{lcccc}
\hline
\textbf{Dataset} & \textbf{Tensor shape} & \textbf{Variables} & $N_t$ \\
\hline
Vallecas (Madrid) & $(8 \times 500 \times 500 \times 30 \times 1)$ & 
$u,\,v,\,w,\,p,\,k,\,\mathrm{CO},\,\mathrm{NO}_x,\,\mathrm{PM}$ & 1  \\
Two-building   & $(3 \times 100 \times 125 \times 50 \times 225)$ & 
$u,\,v,\,w$ & 225  \\
\hline
\end{tabular}
\caption{Summary of the two datasets used in this study.}
\label{tab:datasets}
\end{table}

\section{Results }
\label{sec:results}

The results of the lcSVD and lcHOSVD reconstructions are presented for both datasets. For each case, the number of sensors is first determined, followed by a quantitative assessment of the reconstruction accuracy via the relative root-mean-square error (RRMSE), computed on fluid cells only. The compression (Eqs. (\ref{eq:sensor_ratio}) and (\ref{eq:compression})), representative reconstructed snapshots, and Q-criterion isosurfaces (Eq. (\ref{eq:qcriterion})) are shown to assess the quality of the recovered flow structures.

\subsection{Mode and sensor selection}
\label{sec:results_modes}

The Vallecas domain covers approximately $1705\times1705$~m in the horizontal plane and 70~m in the vertical, discretised with $500\times500\times30$ grid points, about $7.5\times10^{6}$ spatial locations. The retained modes are chosen from the singular-value decay shown in Fig.~\ref{fig:sv_decay} for $k$ and PM, with the remaining variables given in Appendix~\ref{sec:appendix_sv}. For the velocity components and $k$, the $10^{-2}$ threshold is reached at approximately 35-55 modes in the streamwise and normal directions and 11-18 modes in the spanwise direction, so the retained counts are set to $r_x = r_y = 50$, $r_z = 20$, with 30 modes for lcSVD. The spanwise velocity and the pressure exhibit higher small-scale features and more localized flow structures, and so the threshold requires slightly more, and $r_x = r_y = 60$, $r_z = 20$ are retained, with 40 modes for lcSVD. The pollutant spectra behave quite differently, where the $10^{-2}$ crossing is not reached until modes 178-195, reflecting the fine spatial gradients of the road-corridor emission plumes. The retained count is nevertheless kept at 50 modes for these variables. In terms of energy, the contribution of the additional modes is minimal, and extending the basis towards the $10^{-2}$ crossing increases the captured energy only marginally. The sensor requirement, in contrast, grows with the retained modes, since the number of sensors along each axis must reach the number of modes retained along that axis. Matching the crossing would demand close to 180 sensor planes per horizontal direction, a network beyond any realistic monitoring deployment, for a limited gain in accuracy. Retaining 50 modes keeps the pollutants within the same configuration as the velocity fields and preserves the compression.

\begin{figure}[h!]
    \centering
    \begin{subfigure}[b]{0.58\textwidth}
        \centering
        \includegraphics[width=\textwidth]{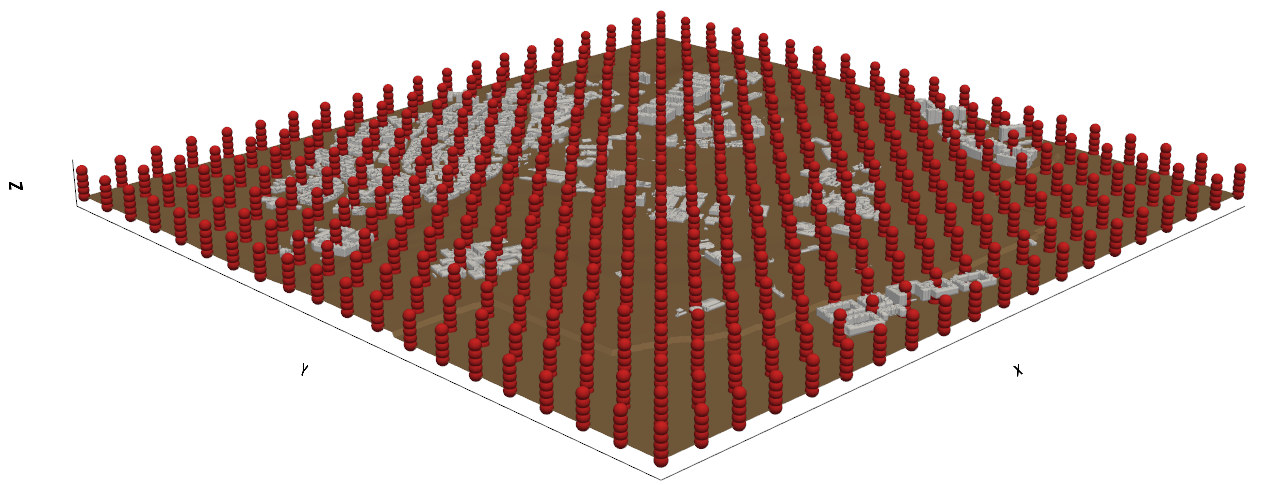}
        \caption{Vallecas urban domain.}
        \label{fig:sensors_vallecas}
    \end{subfigure}
    \hfill
    \begin{subfigure}[b]{0.4\textwidth}
        \centering
        \includegraphics[width=\textwidth]{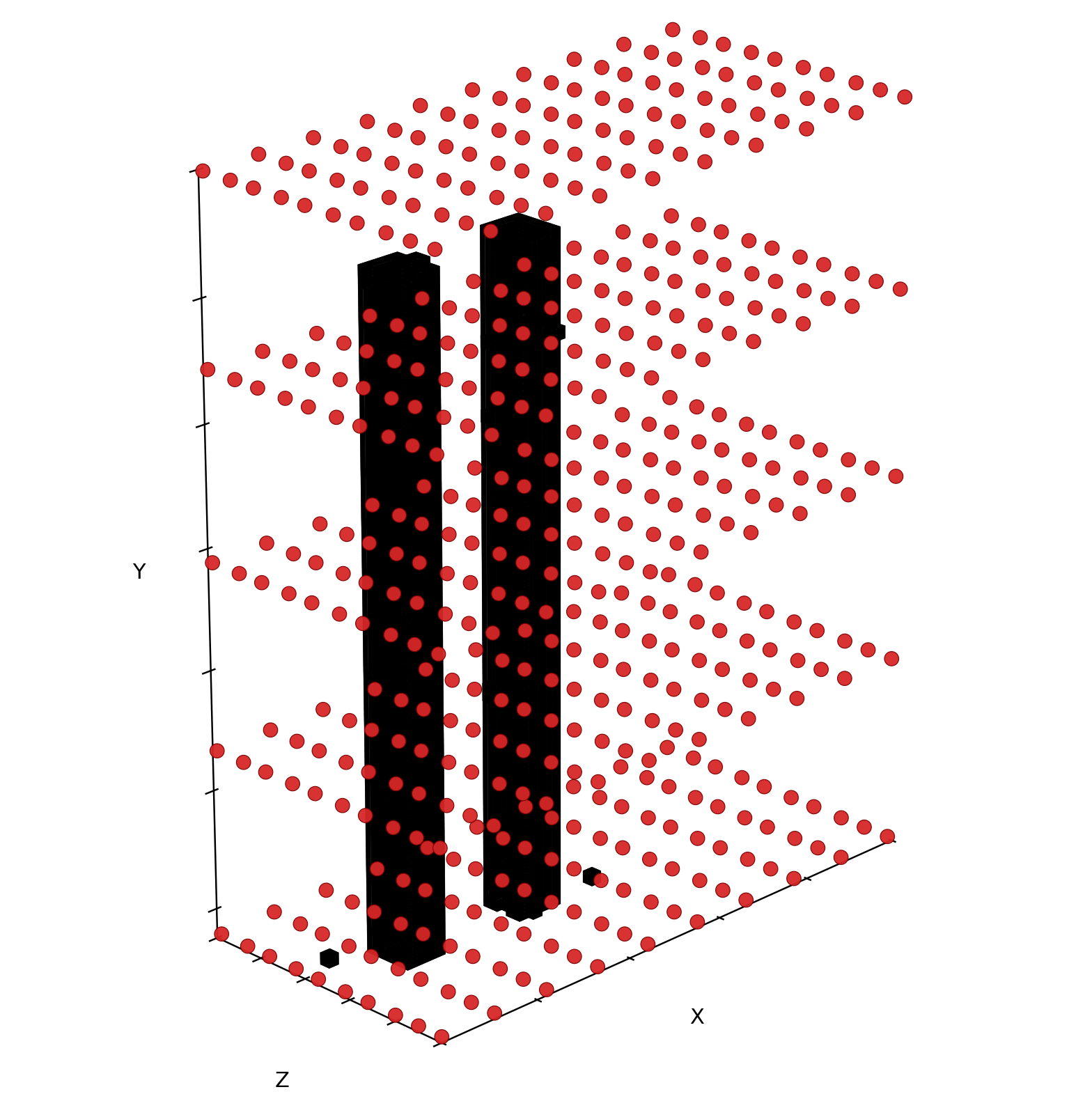}
        \caption{Two-building domain.}
        \label{fig:sensors_twobldg}
    \end{subfigure}
    \caption{Representative sensor distributions employed for the low-cost decompositions. A subset of 2{,}000 (Vallecas) and 500
    (two-building) sensor locations is displayed out of the actual networks of 50{,}000 and 25{,}000 measurement points, respectively.}
    \label{fig:sensor_layouts}
\end{figure}

The sensor networks follow directly from these counts. A network of $50\times50\times20$ sensors is employed for the first group of variables (50{,}000 locations, a horizontal spacing of about 34~m and $CF = 150\times$), and $60\times60\times20$ sensors for the second (72{,}000 locations, about 29~m spacing and $CF = 104\times$). Both configurations use the minimum admissible budget per direction to maximise compression, retaining less than 1\% of the grid points with roughly 1--2 sensors per street block. Figure~\ref{fig:sensor_layouts} illustrates representative subsets of the sensor distributions for both datasets. Only a fraction of the networks is displayed, intended to convey the equidistant arrangement rather than the full density.

\begin{figure}[t]
    \centering
    \begin{subfigure}[b]{0.48\textwidth}
        \centering
        \includegraphics[width=\textwidth]{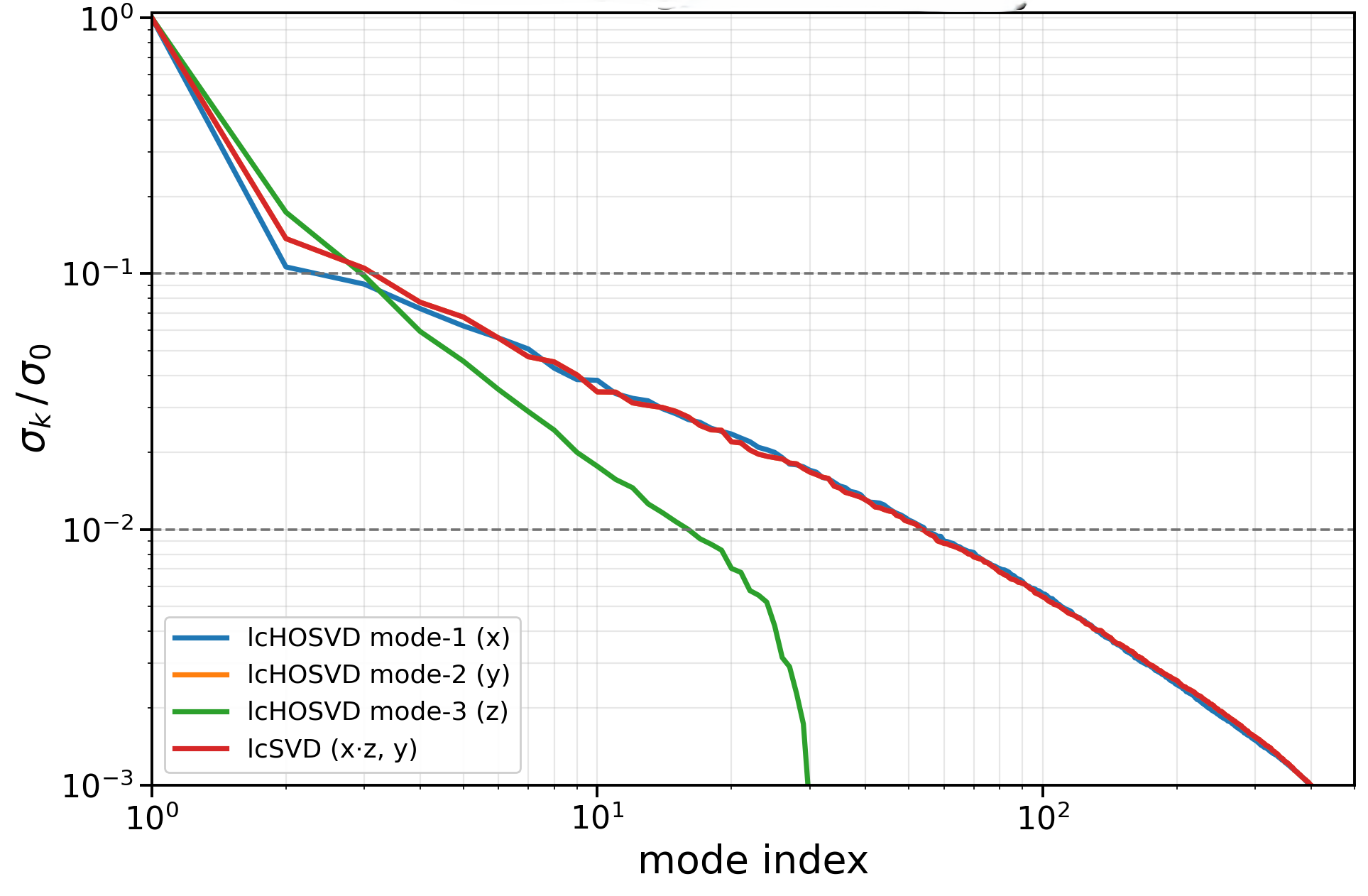}
        \caption{Turbulent kinetic energy ($k$)}
        \label{fig:sv_k}
    \end{subfigure}
    \hfill
    \begin{subfigure}[b]{0.48\textwidth}
        \centering
        \includegraphics[width=\textwidth]{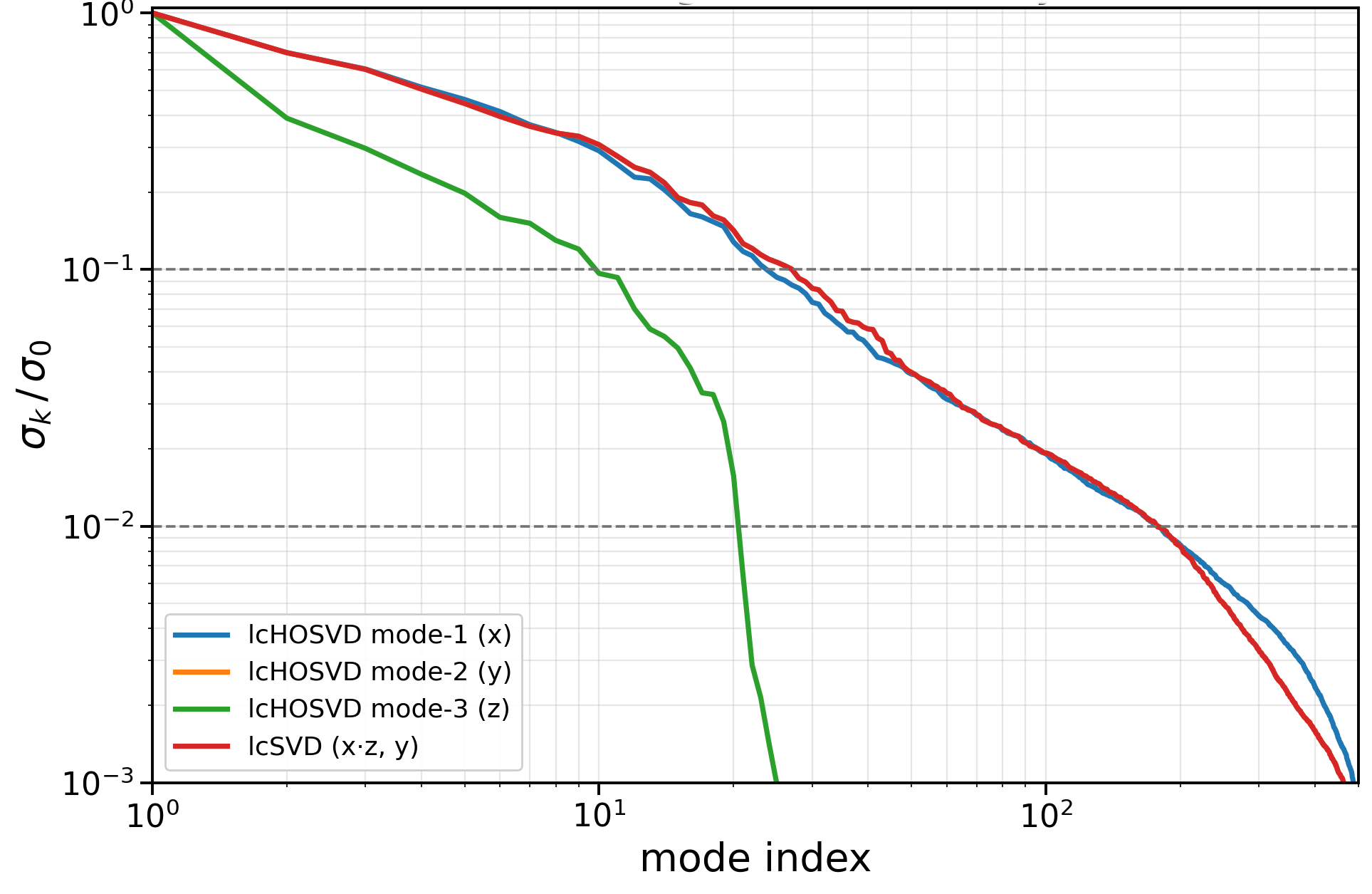}
        \caption{Particulate matter (PM)}
        \label{fig:sv_pm}
    \end{subfigure}
    \caption{Singular-value decay $\sigma_k/\sigma_0$ for the turbulent
    kinetic energy ($k$) and particulate matter (PM) fields in the Vallecas
    dataset.}
    \label{fig:sv_decay}
\end{figure}

\begin{figure}[h!]
    \centering
    \includegraphics[width=\textwidth, height=0.35\textheight]{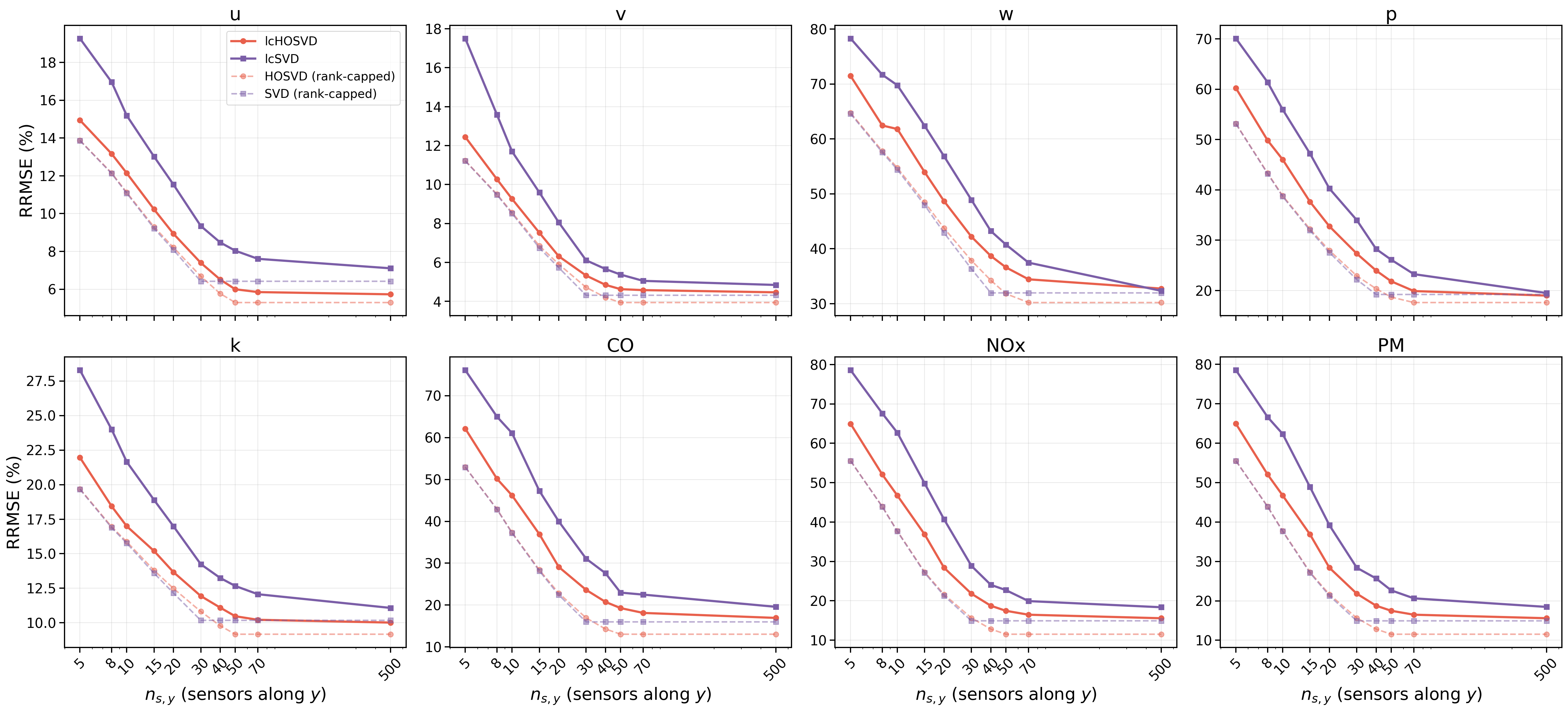}
    \caption{Sensor anisotropy study for the Vallecas case. Relative root-mean-square error (RRMSE) of lcHOSVD and lcSVD as the number of sensor planes in the $y$ direction ($n_{s,y}$) is progressively reduced while keeping the sensor counts along $x$ and $z$ fixed. 50,50 and 20 modes are retained in the $x$,$y$,$z$ directions for the streamwise velocity ($u$), normal velocity ($v$), turbulent kinetic energy ($k$), carbon monoxide (CO), nitrogen oxides (NO$_x$), and particulate matter (PM), and 60,60,20  for the spanwise velocity ($w$) and pressure ($p$), while the lcSVD formulation retains 30 and 40 modes, respectively. The top row presents, from left to right, $u$, $v$, $w$, and $p$. The bottom row presents $k$, CO, NO$_x$, and PM.}
    \label{fig:anisotropy_vallecas}
\end{figure}

The choice of this configuration is supported by an anisotropy analysis, which examines how the reconstruction responds when the sensor count along one direction departs from the nominal value. In practical monitoring campaigns, the sensor distribution is rarely uniform, as access and cost constraints often limit the number of measurement planes along a given spatial direction. To assess the response of both methods to this situation, the reconstruction is repeated with a deliberately anisotropic sensor arrangement. The reconstruction is repeated with the sensor counts along $x$ and $z$ fixed at their nominal values, while the number of sensor planes along $y$, $n_{s,y}$, is reduced progressively. Both methods receive the same total sensor budget at every point of the sweep, and for lcHOSVD only the number of modes along $y$ is capped at $n_{s,y}$. Figure~\ref{fig:anisotropy_vallecas} shows the resulting RRMSE. Adding sensors beyond the nominal configuration brings only marginal gains where increasing $n_{s,y}$ from 50 to full sampling at 500 planes, a tenfold increase in the budget, reduces the lcHOSVD error from 6.00\% to 5.73\% for $u$, from 11.10\% to 10.02\% for $k$, and from 17.45\% to 15.56\% for NO$_x$. Reducing the sensors below the nominal value, in contrast, degrades the reconstruction rapidly, and far more severely for lcSVD. At $n_{s,y}=10$, its error reaches 62.71\% for NO$_x$, 61.07\% for CO, and 62.34\% for PM, against 46.75\%, 46.19\%, and 46.74\% for lcHOSVD. The origin of this behaviour is structural, as in the matrix formulation, the $y$-planes act as the snapshots of the unfolded matrix, so the rank of the lcSVD reconstruction cannot exceed $n_{s,y}$, and starving one direction penalises all three at once. In the tensor formulation, each spatial direction holds its own factor matrix, and the loss of information remains confined to the y modes. The only requirement is that the number of sensor planes reaches the y-rank of the decomposition. As the number of measurement planes $n_{s,y}$ increases, the lcHOSVD error falls steadily and approaches the full HOSVD reference for every variable. This convergence confirms that the larger errors seen at the sparsest sensor configurations come from the limited amount of measurement data rather than from any intrinsic limitation of lcHOSVD, which recovers the full-data baseline once enough planes are sampled. 

\begin{figure}[h!]
    \centering
    \includegraphics[width=\textwidth]{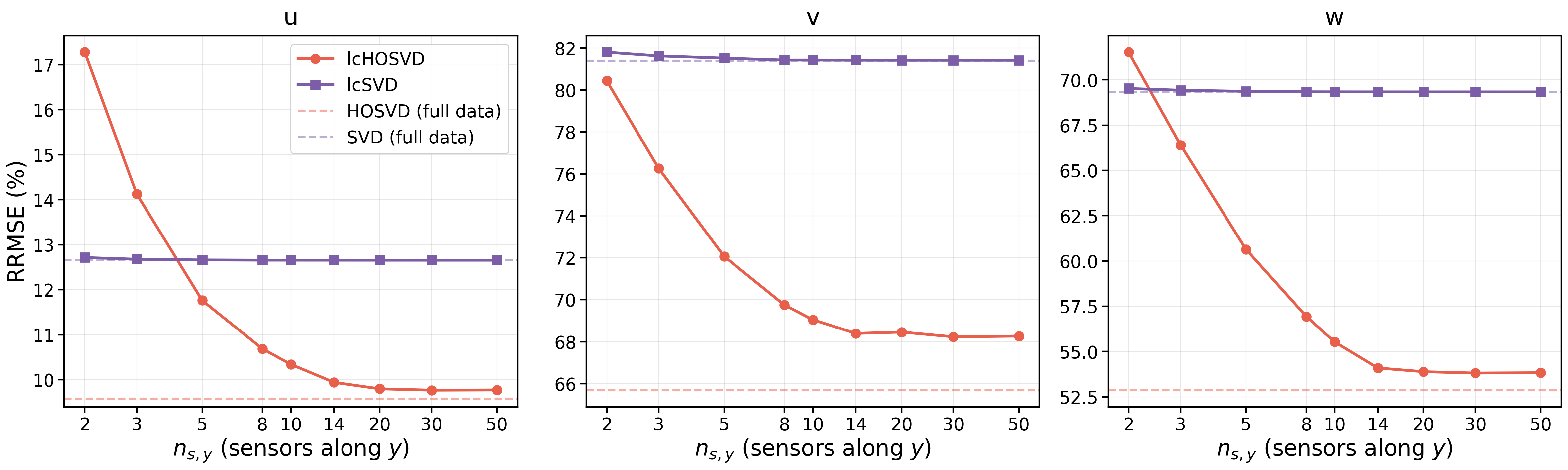}
    \caption{Sensor anisotropy study for the two-building case. Relative root-mean-square error (RRMSE) of lcHOSVD and lcSVD for decreasing numbers of sensor planes in the $y$ direction ($n_{s,y}$). 19,14 and 15 modes are retained along the $x$, $y$, and $z$ directions for all velocity components, while the lcSVD formulation retains 8 modes. From left to right, streamwise velocity ($u$), normal velocity ($v$), and spanwise velocity ($w$).}
    \label{fig:anisotropy_twobldg}
\end{figure}

For the two-building dataset, the $10^{-2}$ threshold of the singular-value decay is reached within 10-20 modes depending on the direction and velocity component, and $r_x = 19, r_y = 14, r_z = 15$ modes are retained for lcHOSVD, covering the most demanding component in each direction, and 8 modes are retained for the lcSVD formulation. The $10^{-1}$ threshold is reached within 3 modes in all directions for every component, reflecting the dominance of large-scale structures in the skimming-flow regime. The sensor configuration is $40\times25\times25$, corresponding to 25{,}000 measurement locations, 4\% of the 625{,}000 grid points ($CF = 25\times$). The same anisotropic sweep is applied to this case, with 40 and 25 sensor planes fixed along $x$ and $z$, and the result is shown in Fig.~\ref{fig:anisotropy_twobldg}. The lcSVD error remains nearly constant, since its rank is fixed by the retained modes rather than by the sensor layout, while the lcHOSVD error decreases with $n_{s,y}$ and settles once the 14 modes along $y$ are matched by the sensor planes, where it lies below the lcSVD error by 2.7 percentage points for $u$, 13.0 for $v$ and up to 15.5 for $w$. For the smallest sensor counts, $n_{s,y} \leq 3$, the $y$ basis of lcHOSVD also collapses, and the advantage disappears for $u$ and $w$. Above this threshold, the tensor method degrades gracefully under anisotropic sampling.

This situation is frequently encountered in practical monitoring campaigns, where sensor placement is constrained by accessibility, infrastructure, and cost considerations. As a result, measurements are often more densely distributed in some spatial directions than in others. By preserving the tensor structure of the data and treating each dimension independently, lcHOSVD can better accommodate such anisotropies, explaining its improved reconstruction performance compared with lcSVD.

\subsection{Qualitative Comparison of Reconstructed Fields}
\label{sec:results_modes}

\begin{figure}[h!]
    \centering

    \includegraphics[
        width=\textwidth,
        height=0.3\textheight,
        keepaspectratio=false
    ]{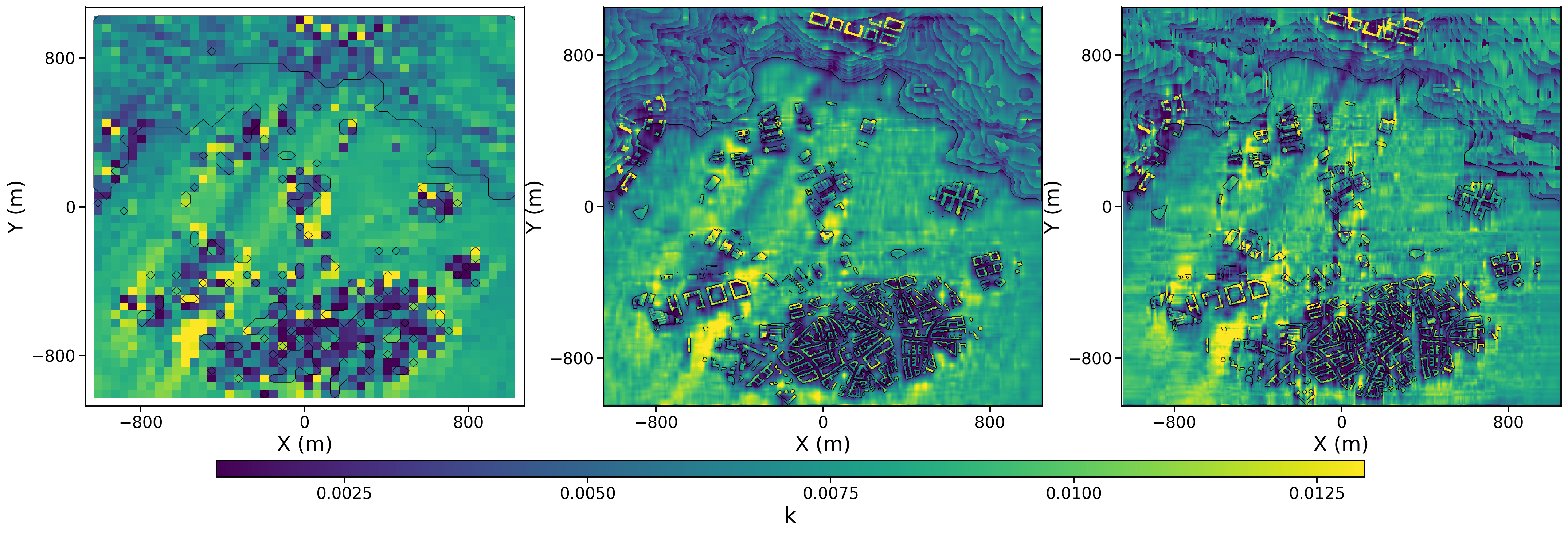}

    \vspace{0.1cm}

    \includegraphics[
        width=\textwidth,
        height=0.3\textheight,
        keepaspectratio=false
    ]{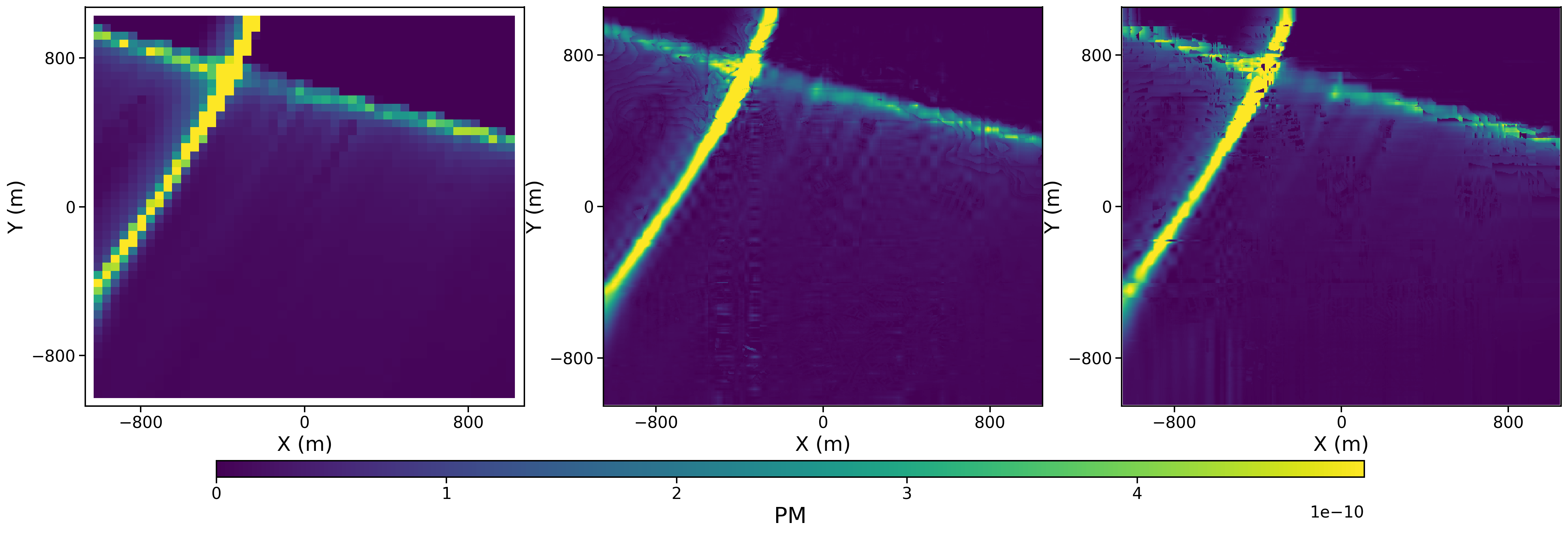}

    \caption{
     From left to right: comparison of the low-cost input, lcHOSVD reconstruction, and lcSVD reconstruction for the turbulent kinetic energy ($k$) and particulate matter (PM) fields in the Vallecas urban-flow dataset, corresponding to 50{,}000 sensor locations out of the $7.5\times10^{6}$ grid points ($CF = 150\times$), with $(50, 50, 20)$ modes for lcHOSVD and 30 modes for lcSVD. The lines mark where terrain or buildings rise above the z = 5 m AGL plane.
    }
    \label{fig:urban_lc_compare}
\end{figure}

The reconstructed snapshots for the Vallecas dataset are presented in Fig.~\ref{fig:urban_lc_compare} for the $k$ and \( \mathrm{PM} \) fields, while the remaining variables are provided in Appendix~\ref {sec:app}. In both cases, the low-cost input contains only a sparse representation of the original field, requiring the dominant spatial structures to be recovered through the reduced-order reconstruction. Both methods reconstruct the large-scale distribution of the variables, reproducing the principal gradients and features observed in the reference solution.

LcHOSVD captures the dominant structures of \( k \) and \( \mathrm{PM} \) slightly better and produces smoother spatial fields. Both methods exhibit reconstruction artefacts arising from the sparse spatial sampling; however, these artefacts appear less pronounced in the lcHOSVD reconstructions. The lcSVD results display a scattered appearance, particularly in regions of strong spatial variation. Nevertheless, the overall agreement between the two methods remains high, confirming that both approaches recover the dominant behaviour of the urban-flow fields despite the substantial spatial compression employed.

\begin{figure}[h!]
    \centering
    \includegraphics[width=\textwidth]{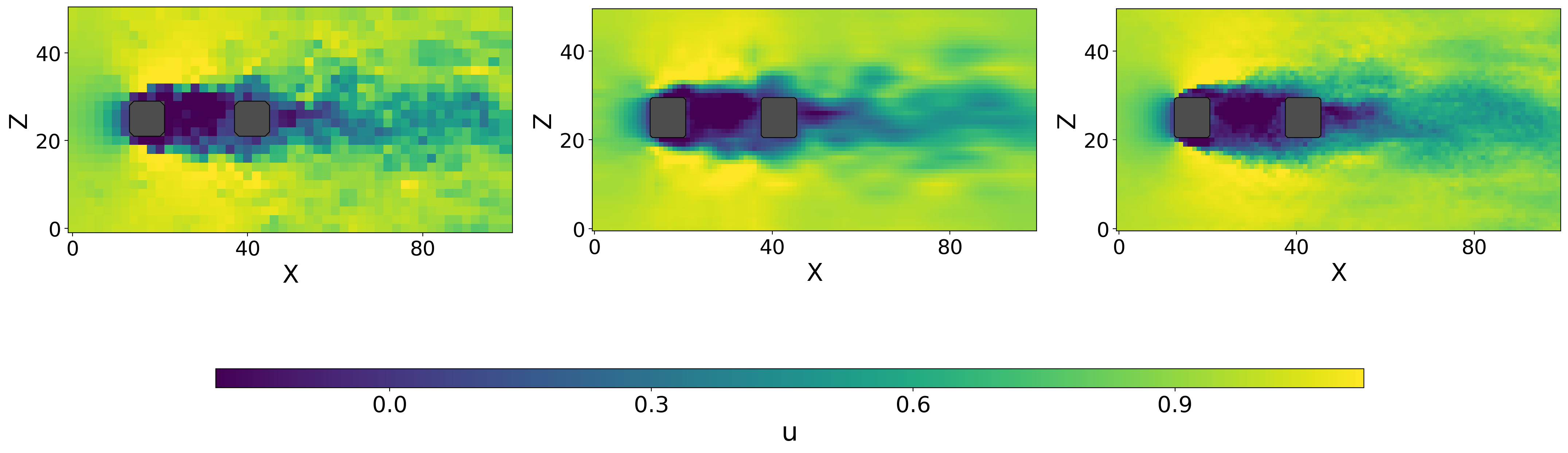}
    \vspace{0.2cm}

    \includegraphics[width=\textwidth]{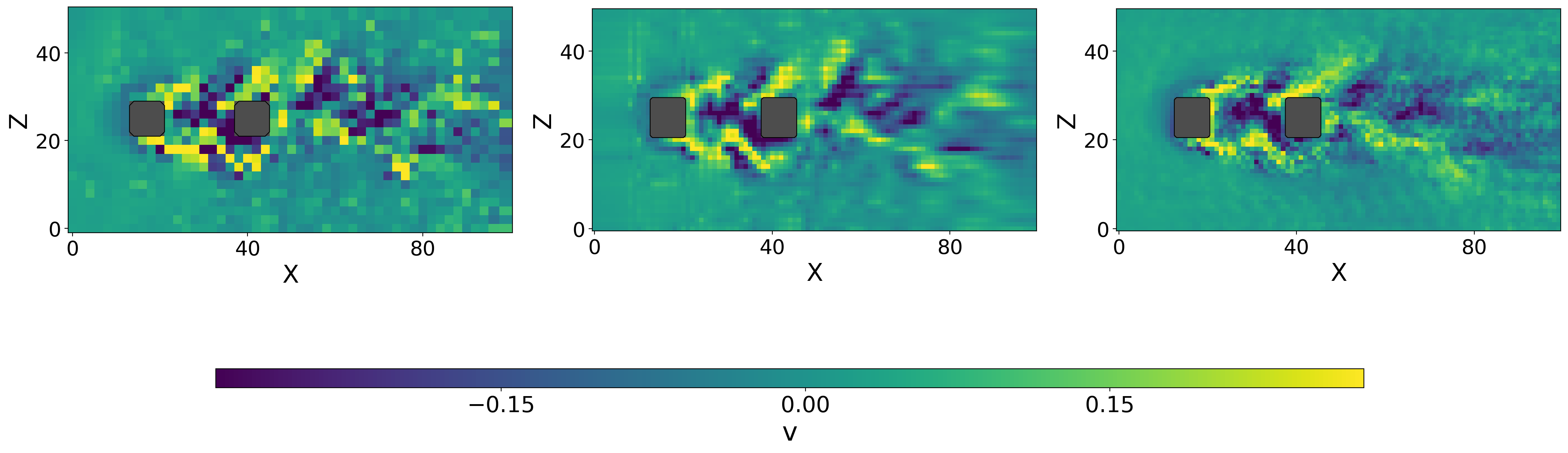}
    \vspace{0.2cm}

    \includegraphics[width=\textwidth]{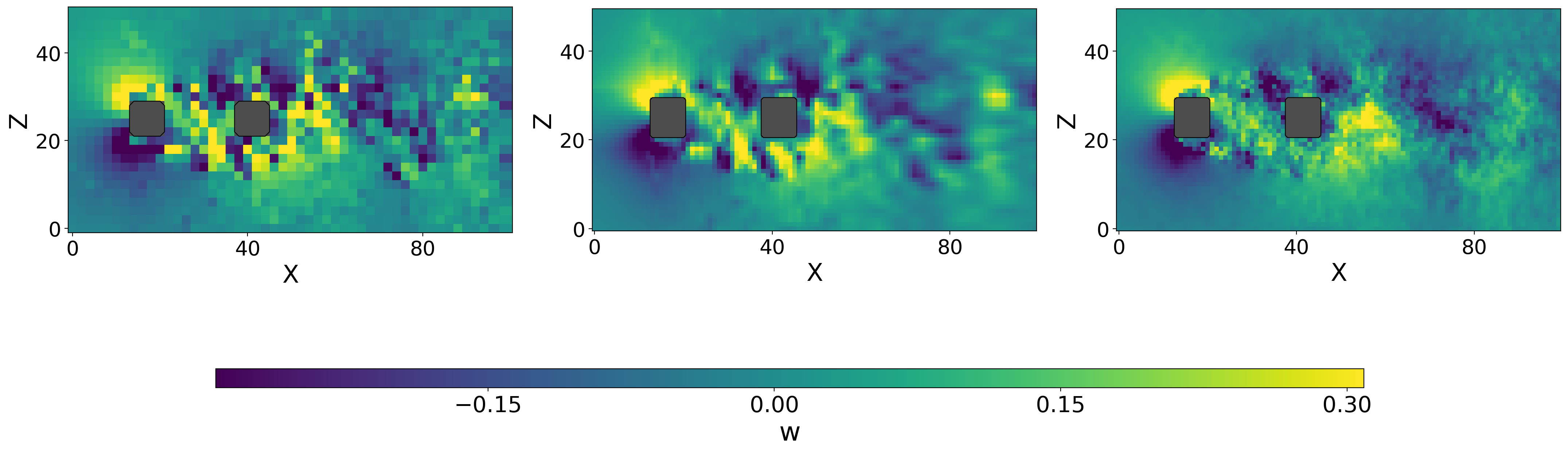}

    \caption{
    From left to right: comparison of the low-cost input, lcHOSVD reconstruction, and lcSVD reconstruction. from top to bottom: streamwise ($u$), wall-normal ($v$), and spanwise ($w$) velocity components at $t=224$ in the two-building dataset, corresponding to 25{,}000 sensor locations out of the 625{,}000 grid points ($CF = 25\times$), with $(19, 14, 15)$ modes for lcHOSVD and 8 modes for lcSVD.
    }
    \label{fig:twobldg_lc_compare}
\end{figure}

The corresponding results for the two-building dataset are shown in Fig.~\ref{fig:twobldg_lc_compare}. For $u$, both lcHOSVD and lcSVD reproduce the wake deficit and the downstream recovery of the flow with comparable accuracy. Differences become more evident in the transverse velocity components. In the normal velocity field, lcHOSVD provides a clearer reconstruction of the alternating flow structures generated by the interaction of the obstacle wakes, whereas the lcSVD reconstruction is more diffused. This behaviour is even more apparent for $w$, where lcHOSVD preserves a greater portion of the coherent structures and sharper gradients visible in the reference field. Both methods recover the dominant flow patterns successfully, although the preservation of the tensor structure in lcHOSVD contributes to improved reconstruction of more complex turbulent flows.

\subsection{Quantitative Comparison of Reconstruction Accuracy and Speed-Up}

The reconstruction accuracy and computational performance of the low-cost decompositions are reported in Tables~\ref{tab:vallecas_rrmse_speedup} and~\ref{tab:twobldg_rrmse_speedup}. For the Vallecas dataset, the results are reported for all variables, including the standard HOSVD reference, lcHOSVD, and lcSVD. For the two-building dataset, the comparison is restricted to lcHOSVD and lcSVD, and the error is reported relative to the HOSVD reconstruction. This is appropriate for the turbulent case, where the HOSVD reconstruction represents the reduced-order reference solution.

\begin{table}[h!]
\centering
\resizebox{\textwidth}{!}{
\begin{tabular}{c|cc|ccc|ccc}
\hline
Component
& \multicolumn{2}{c|}{HOSVD}
& \multicolumn{3}{c|}{lcHOSVD}
& \multicolumn{3}{c}{lcSVD} \\
\cline{2-9}
& RRMSE (\%) & $t$ (s)
& RRMSE (\%) & $t$ (s) & Speed-up
& RRMSE (\%) & $t$ (s) & Speed-up \\
\hline
$u$    & 5.30  & 1.758 & 6.00  & 0.458 & 3.84 & 8.04  & 0.024 & 72.80 \\
$v$    & 3.94  & 1.721 & 4.63  & 0.455 & 3.78 & 5.38  & 0.026 & 65.70 \\
$w$    & 30.21 & 1.868 & 34.91 & 0.497 & 3.76 & 39.22 & 0.036 & 51.86 \\
$p$    & 17.61 & 1.816 & 20.79 & 0.498 & 3.64 & 24.23 & 0.035 & 51.46 \\
$k$    & 9.18  & 1.809 & 11.10 & 0.471 & 3.84 & 13.25 & 0.025 & 71.61 \\
CO     & 13.04 & 1.831 & 19.28 & 0.463 & 3.96 & 22.98 & 0.025 & 72.13 \\
NO$_x$ & 11.50 & 1.763 & 17.45 & 0.479 & 3.68 & 22.71 & 0.025 & 69.27 \\
PM     & 11.53 & 1.807 & 17.48 & 0.489 & 3.70 & 22.66 & 0.026 & 70.83 \\
\hline
\end{tabular}
}
\caption{Reconstruction error and timing for HOSVD, lcHOSVD, and lcSVD on the Vallecas case.}
\label{tab:vallecas_rrmse_speedup}
\end{table}

\begin{table}[h!]
\centering
\begin{tabular}{c|ccc|ccc}
\hline
\multirow{2}{*}{Component}
& \multicolumn{3}{c|}{lcHOSVD}
& \multicolumn{3}{c}{lcSVD} \\
\cline{2-7}
& $\Delta$RRMSE & $t$ (s) & Speed-up
& $\Delta$RRMSE & $t$ (s) & Speed-up \\
\hline
$u$ & 0.28 & 23.10 & 2.1$\times$ & 3.08  & 1.32 & 37.2$\times$ \\
$v$ & 2.72 & 22.30 & 2.2$\times$ & 15.73 & 1.36 & 36.2$\times$ \\
$w$ & 1.35 & 22.44 & 2.2$\times$ & 16.47 & 1.60 & 30.6$\times$ \\
\hline
\end{tabular}
\caption{RRMSE, computational time, and speed-up for the two-building dataset.}
\label{tab:twobldg_rrmse_speedup}
\end{table}

A clear trade-off is observed between reconstruction accuracy and computational efficiency. In both datasets, lcSVD provides substantially larger computational savings, whereas lcHOSVD consistently achieves lower reconstruction errors by preserving the multi-dimensional structure of the original data during the decomposition process. The difference in reconstruction accuracy for the urban flow case between the two approaches remains relatively modest. The largest discrepancy is observed for the PM field, where the RRMSE (Eq. (\ref{eq:rrmse})) increases from 17.48\% for lcHOSVD to 22.66\% for lcSVD, while for the $k$ field the increase is from 11.10\% to 13.25\%. The comparatively high RRMSE values obtained for the \(w\), \(p\), and pollutant fields are a consequence of the low magnitude of these variables over large portions of the domain. As the RRMSE is divided by zero or near-zero values, it leads to disproportionately large relative errors, thereby inflating the overall RRMSE between the reconstructed and reference fields. Despite these differences, lcSVD provides a substantial computational advantage. Depending on the variable considered, the speed-up increases from approximately 51$\times$ to 72$\times$ for lcSVD, compared with only 3.6$\times$ to 4.0$\times$ for lcHOSVD. Given that the urban-flow dataset contains a single temporal step, the increase in reconstruction error remains relatively small compared with the significant reduction in computational cost, making lcSVD an attractive option when rapid reconstruction is the primary objective.

The two-building case shows a different balance between the two methods. With respect to the standard HOSVD reference, lcHOSVD increases the RRMSE by only 0.28 percentage points for $u$, 2.72 for $v$, and 1.35 for $w$, while lcSVD adds 3.08, 15.73, and 16.47 points, respectively. In the direct comparison, lcHOSVD is therefore more accurate than lcSVD by 2.80, 13.01, and 15.12 percentage points, while lcSVD remains 14-18 times faster (mean runtimes of 1.42~s against 22.62~s, corresponding to speed-ups of 34$\times$ and 2.2$\times$ with respect to HOSVD). The larger accuracy gap for the $v$ and $w$ components indicates that the tensor formulation gains importance as the directional anisotropy of the flow increases. Although lcSVD retains a clear computational advantage, the better reconstruction obtained with lcHOSVD for $v$ and $w$ shows that preserving the multi-way structure of the data is beneficial when representing the wake dynamics characteristic of turbulent building flows.

The difference between the two methods can also be quantified through the storage cost in the reduced representation. For the Vallecas dataset, the core tensor combines the directional modes, so the entire storage spans $50\times50\times20$ admissible mode combinations, whereas the 30 lcSVD modes provide exactly 30 global spatial patterns. In other words, the tensor representation retains a substantially larger set of admissible reduced-order interactions as compared to lcSVD. This bigger set of modal interactions allows lcHOSVD to capture more complex spatial structures and directional dependencies, which contribute to its lower reconstruction errors. The computational cost follows the opposite logic, where lcSVD computes a single SVD of one small subsampled matrix, while lcHOSVD requires one SVD per spatial direction plus the core requirements, which is why the lcSVD speed-ups exceed those of lcHOSVD despite its smaller parameter count.

\subsection{Q-criterion}

\begin{figure}[h!]
    \centering

    \begin{subfigure}[b]{1\textwidth}
        \centering
        \includegraphics[width=\textwidth]{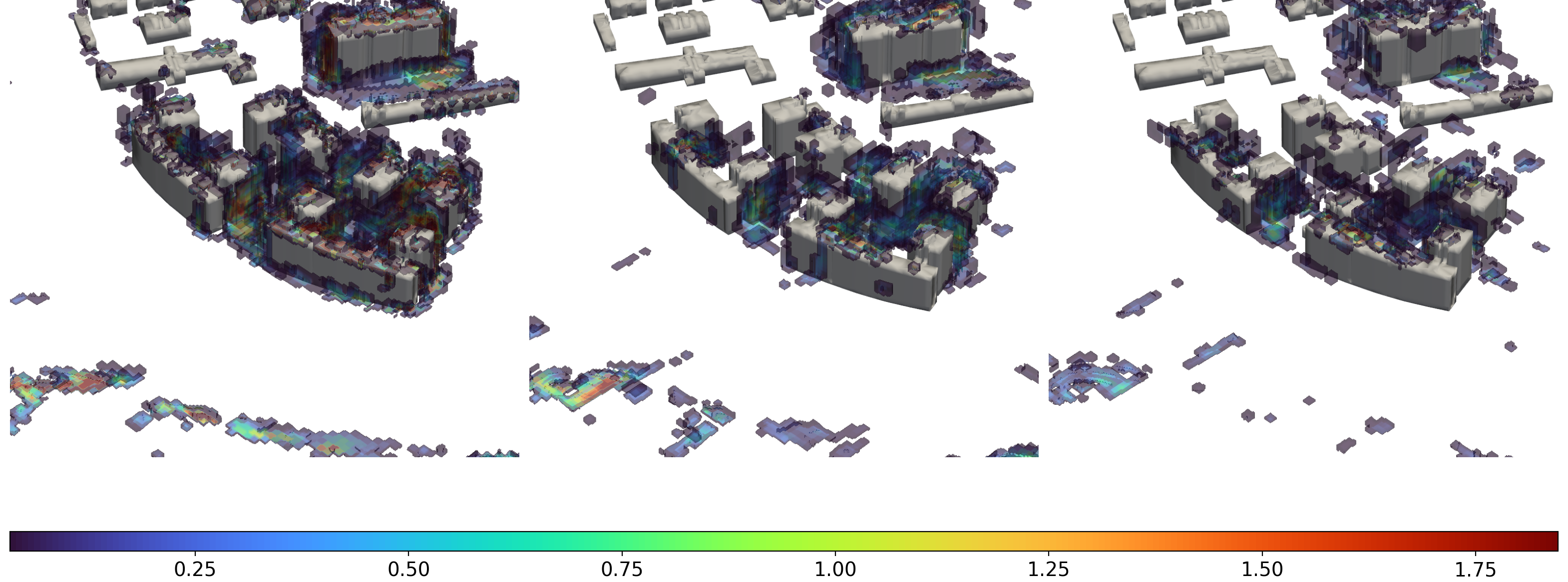}
        
    \end{subfigure}

    \vspace{0.4cm}

    \begin{subfigure}[b]{1\textwidth}
        \centering
        \includegraphics[width=\textwidth]{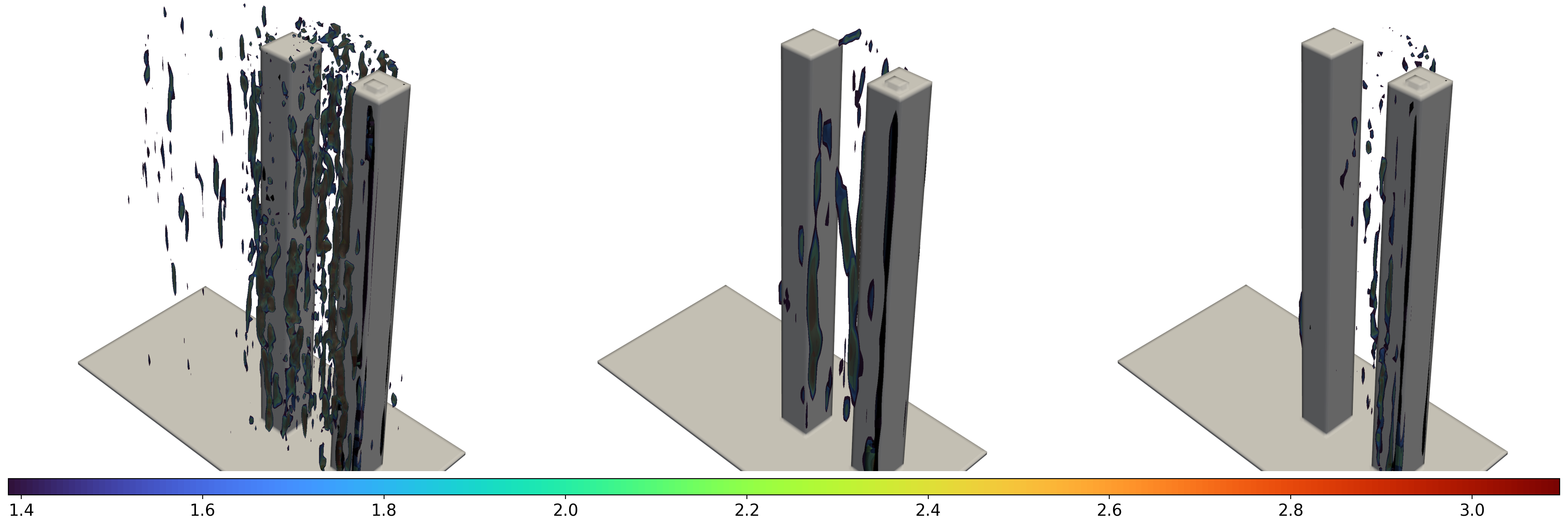}
        
    \end{subfigure}

    \caption{
    From left to right: Comparison of standardized Q-criterion isosurfaces for the Vallecas dataset(top) and the two-building dataset (bottom) obtained from the ground-truth, lcHOSVD reconstruction, and lcSVD reconstruction, respectively.}
    \label{fig:qcriterion_vallecas}
\end{figure}

Figure~\ref{fig:qcriterion_vallecas} compares the standardized Q-criterion isosurfaces (Eq. (\ref{eq:qcriterion})) obtained from the reference solution, lcHOSVD reconstruction, and lcSVD reconstruction. The Q-criterion is standardized using the statistics of the reference field and visualized using identical thresholds and colour scales. A zoomed view of the Vallecas domain is presented to improve visual clarity and allow a more detailed inspection of the coherent vortical structures present within the urban canopy. The Q-criterion comparison suggests that both reconstruction methods capture the dominant large-scale vortical structures present in the reference solution. However, the reconstructed fields appear to contain fewer Q-criterion isosurfaces than the original CFD data, indicating a partial loss of smaller-scale coherent structures. This effect seems to be less pronounced for lcHOSVD, which preserves a larger fraction of the vortical features visible in the reference field. Although lcHOSVD captures more features than lcSVD, both approaches provide a comparable representation of the large-scale coherent motions present in the urban flow.

The behaviour is more clearly differentiated in the two-building case, where the ground truth contains more complex turbulent structures. The reconstructed Q-criterion isosurfaces show that the lcHOSVD reconstruction preserves the shape, continuity, and spatial distribution of these vortical structures more accurately, while capturing the larger features. The lcSVD reconstruction exhibits a greater degree of fragmentation and loss of features. This observation agrees with the reconstruction-error analysis of the velocity components, particularly for the transverse and vertical velocities, where the tensor-based formulation outperformed the matrix-based approach along the normal and spanwise directions. The results indicate that the advantages of preserving the multidimensional structure of the data become increasingly important as the flow complexity and turbulence content increase.

\begin{figure}[h!]
    \centering
    \includegraphics[width=0.5\textwidth]{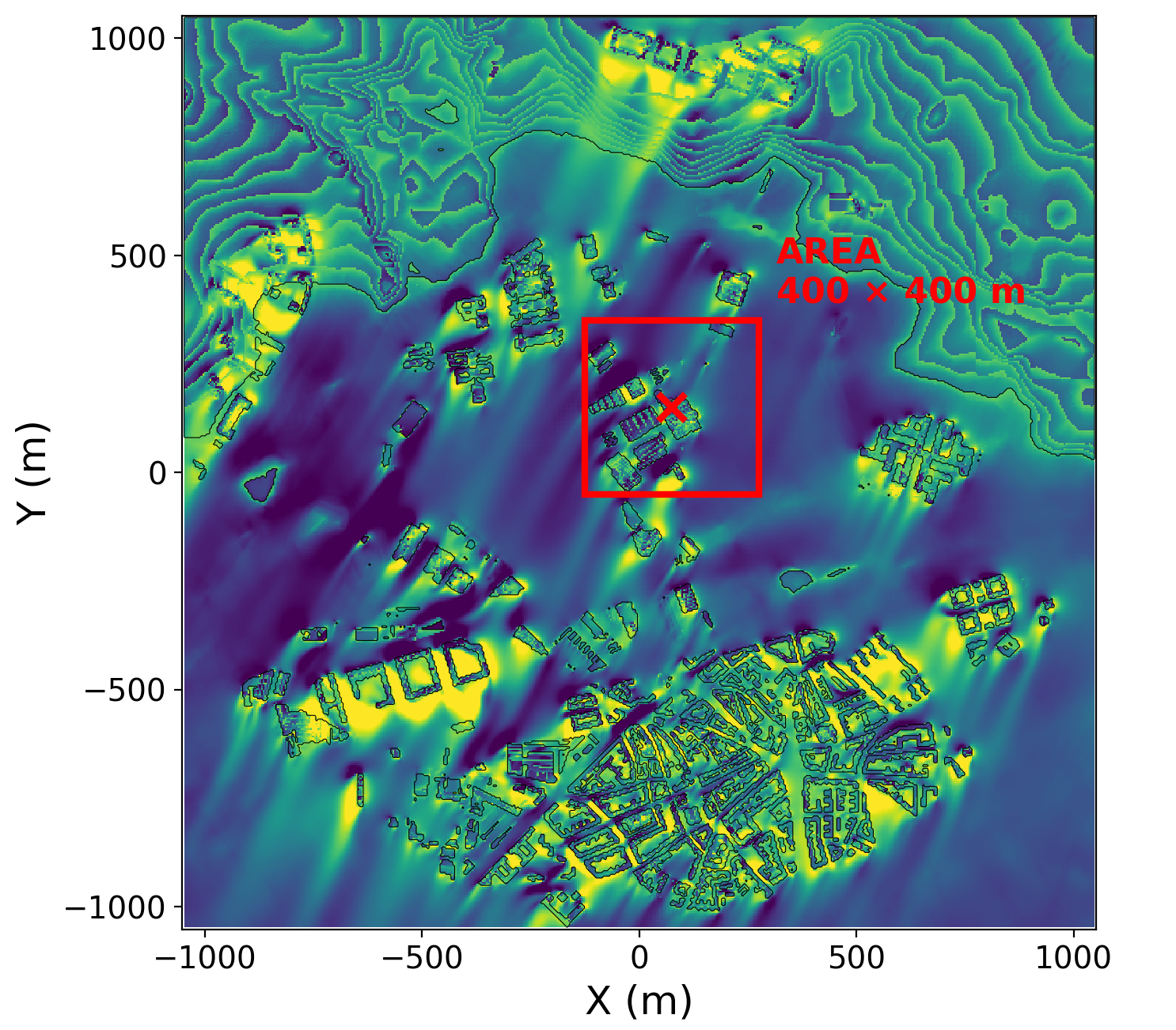}
    \caption{Location of the 400 $\times$ 400m  area (red box) within the Vallecas domain, centred at $(X, Y) = (75, 150)$ m.}
    \label{fig:roi_location_vallecas}
\end{figure}

\begin{figure}[h!]
    \centering
    \includegraphics[width=\textwidth]{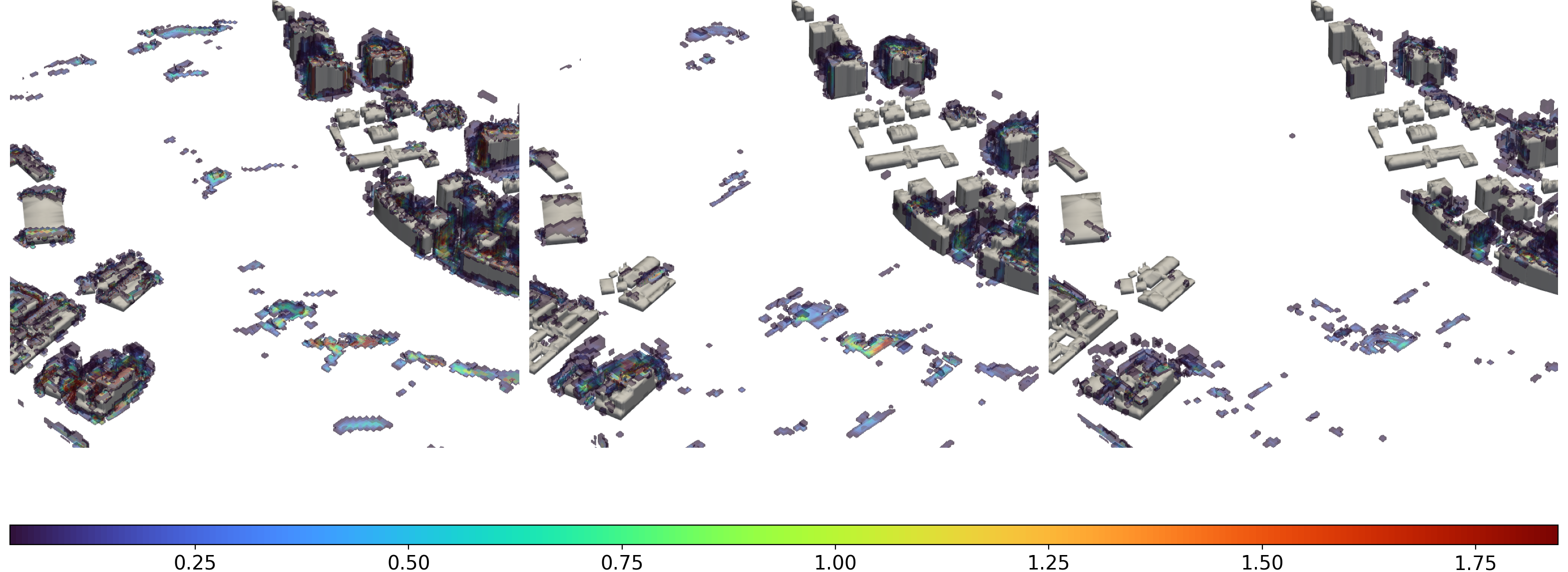}
    \caption{ Q-criterion isosurfaces within the 400~$\times$~400m~ region. From left to right: ground-truth, lcHOSVD reconstruction, and lcSVD reconstruction.}
    \label{fig:appendix_qcriterion}
\end{figure}

From a practical standpoint, the subdomain shown in Fig.~\ref{fig:appendix_qcriterion}, extracted from the urban domain presented in Fig.~\ref{fig:roi_location_vallecas}, suggests that an urban area of approximately 400 m × 400 m can be adequately characterized using around 9 sensors along each horizontal direction and 20 sensors along the vertical direction. This corresponds to one measurement every 44 m in the X-Y plane. This sampling density is feasible in real urban deployments at a local scale. Low-cost air quality and wind sensors mounted on lamp posts, traffic lights, and rooftops, complemented by a small number of vertical profiles from sodars or drone-based surveys, can supply the input fields required by the reduced-order model without dense instrumentation. The vertical resolution is the most demanding part of the requirement, but it can be relaxed in operational settings by placing measurements at carefully chosen heights, since the dominant flow features and pollutant gradients are concentrated near the surface and within the urban canopy. The combination of sparse spatial sampling and modal reconstruction makes the approach suitable for city-scale wind and air quality monitoring, where dense reference data is rarely available.


\section{Conclusions}
\label{sec:conclusions}

This study introduced low-cost variants of SVD and HOSVD for the reconstruction of high-dimensional fluid-flow and urban air-quality datasets from sparse sensor measurements. The proposed methodology performs the decomposition using spatially reduced observations and subsequently reconstructs the full-resolution fields, enabling accurate recovery of complex flow structures while significantly reducing the computational burden associated with conventional modal decompositions.

The approach was assessed using datasets with distinct dynamical characteristics. For the urban-flow and pollutant-transport case, both lcSVD and lcHOSVD successfully reproduced the dominant flow and concentration patterns using only a small fraction of the available spatial locations. Reconstruction differences between the two formulations remained relatively modest, with lcHOSVD providing slightly cleaner reconstructions and lcSVD achieving substantially larger computational speed-ups.

Different behaviour was observed for the two-building configuration. Although lcSVD remained considerably faster, reconstruction errors increased more noticeably for the transverse and vertical velocity components. The advantages of the tensor-based formulation became particularly evident in the recovery of coherent vortical structures and small-scale turbulent dynamics, where lcHOSVD consistently produced more accurate reconstructions. These results demonstrate that preserving the multidimensional structure of the data becomes increasingly important as the complexity, unsteadiness, and coupling of the underlying dynamics increase.

The sensor-anisotropy study further demonstrated that the advantages of lcHOSVD extend beyond reconstruction accuracy alone. While lcSVD is fundamentally constrained by the most sparsely sampled direction, lcHOSVD degrades more gracefully because each spatial direction is represented independently through its own factor matrix. This property makes the tensor formulation more robust to anisotropic sampling and non-uniform sensor distributions, situations that commonly arise in practical monitoring networks due to physical, logistical, and economic constraints.

The principal contribution of this work is the introduction of lcHOSVD, a novel sparse-sensing tensor-reconstruction framework that extends low-cost reduced-order modelling of multidimensional datasets beyond matrix-based formulations. To the authors’ knowledge, this is the first methodology capable of performing HOSVD-based field reconstruction directly from sparse sensor observations while retaining the advantages of tensor representations. By preserving the natural multidimensional organisation of the data, lcHOSVD exploits correlations across spatial, temporal, and physical-variable dimensions that cannot be fully captured using matrix-based approaches.

This work also presents the first application of low-cost SVD- and HOSVD-based reconstruction techniques to urban flow and air-quality datasets. The comparative analysis provides practical guidelines for selecting between matrix- and tensor-based formulations. When computational speed is the primary requirement, and the underlying dynamics remain relatively simple, lcSVD offers an attractive solution. Conversely, lcHOSVD provides a more robust framework for strongly unsteady and multidimensional problems, particularly when different spatial directions exhibit distinct levels of complexity or sensor availability.

Beyond reconstruction, the proposed methodologies offer significant potential for data assimilation, digital twins, environmental monitoring, and forecasting applications, where complete flow and concentration fields must be estimated from limited observations. The ability to recover accurate high-dimensional states from sparse sensor networks opens new opportunities for integrating experimental measurements with large-scale numerical simulations in urban and environmental systems.

Future work will focus on incorporating the proposed framework into data-assimilation and digital-twin environments for urban-flow and air-quality forecasting. Additional developments will investigate low-cost tensor-based modal decomposition techniques, including a low-cost Higher-Order Dynamic Mode Decomposition (lcHODMD), adaptive sensor-placement strategies, and the use of low-cost modal representations as compressed inputs for deep-learning models. Further applications to larger multi-physics and environmental datasets will also be explored.

\section*{Conflicts of Interest} 
The authors declare no conflict of interest.


\section*{Code and Data Availability}
The code and data developed for this study are available at:  
\noindent\url{https://modelflows.github.io/modelflowsapp/}.
\section*{Acknowledgments}
The authors acknowledge the MODELAIR project that has received funding from the European Union’s Horizon Europe research and innovation programme under the Marie Sklodowska-Curie grant agreement No. 101072559. The results of this publication reflect only the author's view and do not necessarily reflect those of the European Union. The European Union cannot be held responsible for them. S.L.C. acknowledges the grant PID2023-147790OB-I00 funded by MCIU/AEI/10.13039 /501100011033 /FEDER, UE. The authors gratefully acknowledge the Universidad Politécnica de Madrid (www.upm.es) for providing computing resources on the Magerit Supercomputer.


\renewcommand\theequation{\Alph{section}\arabic{equation}} 
\counterwithin*{equation}{section} 
\renewcommand\thefigure{\Alph{section}\arabic{figure}} 
\counterwithin*{figure}{section} 
\renewcommand\thetable{\Alph{section}\arabic{table}} 
\counterwithin*{table}{section} 

\begin{appendices}

\section{Appendix}
\label{sec:app}
The remaining reconstruction and singular value decay results for the Vallecas and the two-building urban-flow dataset are presented in the section below.

\subsection{Additional Reconstruction Results for the Vallecas Dataset}

\begin{figure}[h!]
\centering

\includegraphics[width=\textwidth]{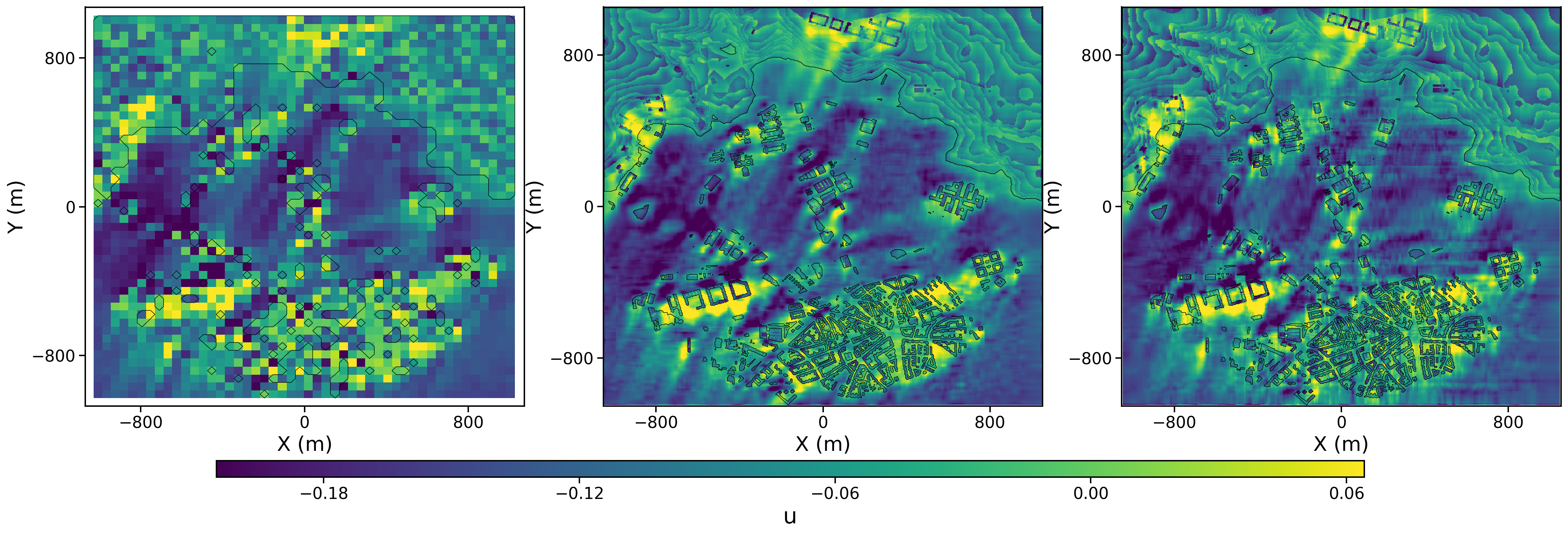}

\vspace{0.05cm}

\includegraphics[width=\textwidth]{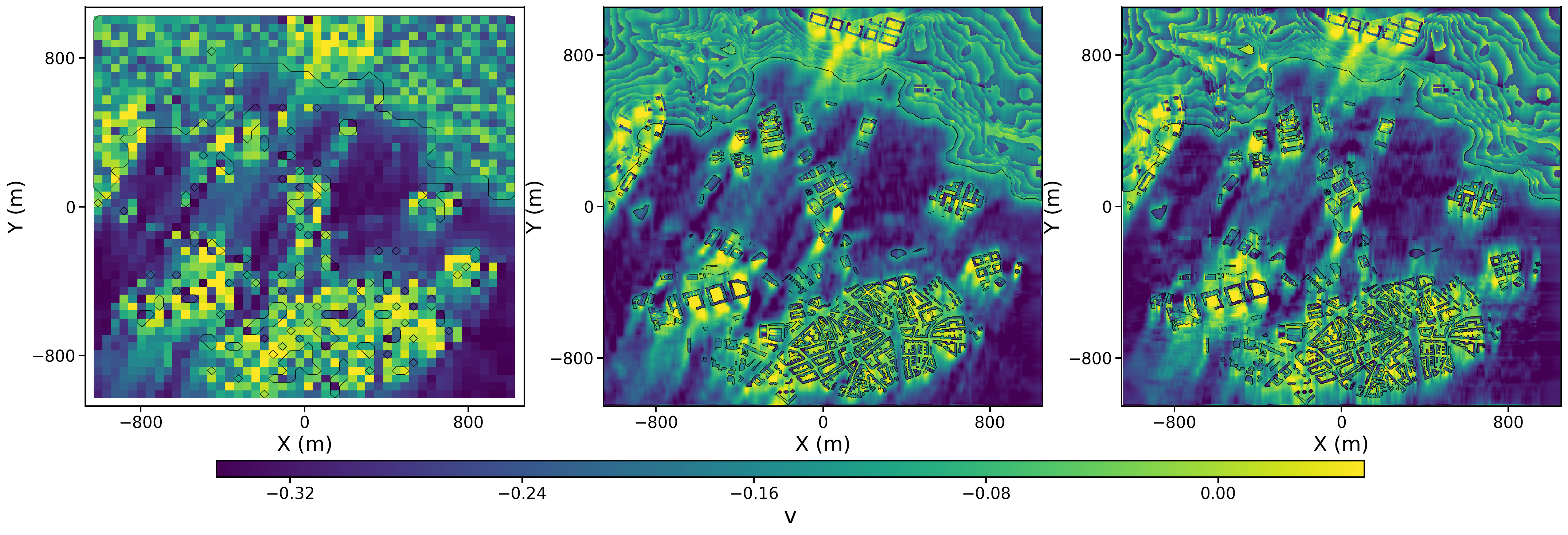}

\vspace{0.05cm}

\includegraphics[width=\textwidth]{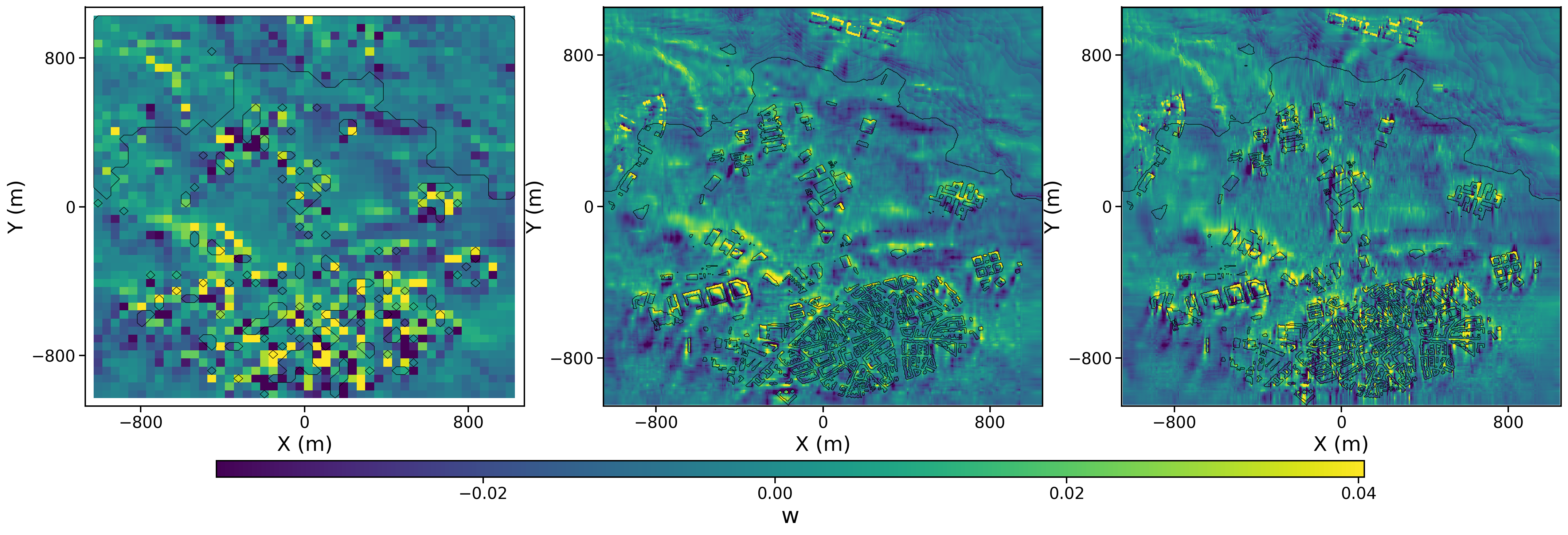}

\vspace{0.05cm}

\includegraphics[width=\textwidth]{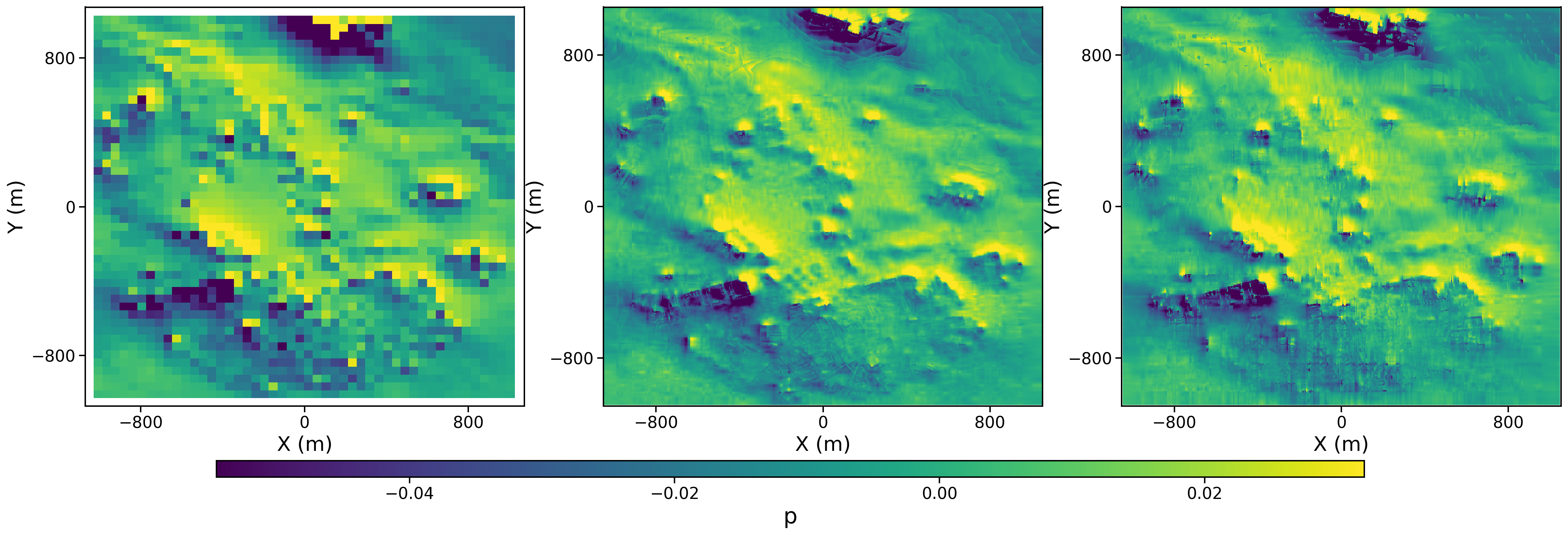}

\caption{
From left to right: low-cost input, lcHOSVD reconstruction, and lcSVD reconstruction for the physical flow variables in the Vallecas dataset, corresponding to 50{,}000 sensor locations out of the $7.5\times10^{6}$ grid points ($CF = 150\times$), with $(50, 50, 20)$ modes for lcHOSVD and 30 modes for lcSVD. From top to bottom: streamwise velocity, normal velocity, spanwise velocity, and pressure. The lines mark where terrain or buildings rise above the $z=5$~m AGL plane.}

\label{fig:appendix_physical}

\end{figure}

\begin{figure}[h!]
\centering

\includegraphics[width=\textwidth]{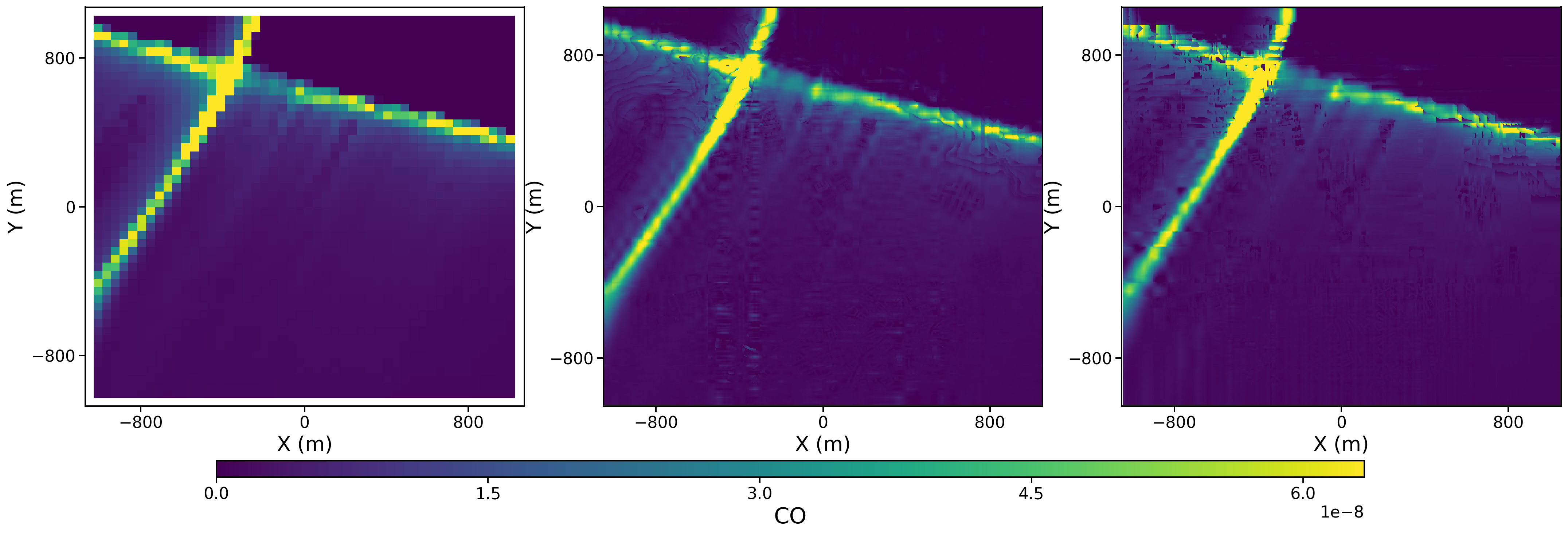}

\vspace{0.05cm}

\includegraphics[width=\textwidth]{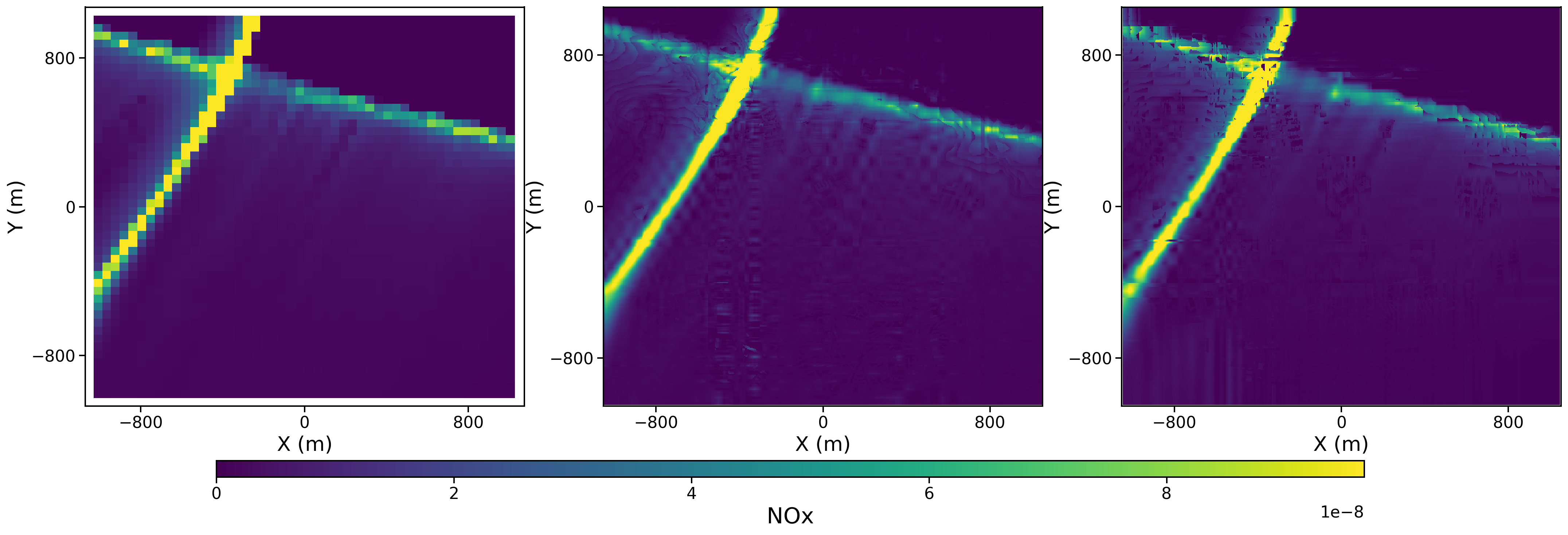}

\caption{
From left to right: low-cost input, lcHOSVD reconstruction, and lcSVD reconstruction for the CO and NOx pollutant fields in the Vallecas dataset, corresponding to 50{,}000 sensor locations out of the $7.5\times10^{6}$ grid points ($CF = 150\times$), with $(50, 50, 20)$ modes for lcHOSVD and 30 modes for lcSVD. From top to bottom: CO and NOx.}

\label{fig:appendix_pollutants}
\end{figure}

The figures include comparisons of the low-cost input, lcHOSVD reconstruction, and lcSVD reconstruction for the velocity components ($u$, $v$, and $w$), pressure ($p$), carbon monoxide (CO), and nitrogen oxides (NO$_x$). These variables exhibit trends similar to those discussed in the main text, where both low-cost approaches recover the dominant flow and pollutant transport patterns. The corresponding reconstruction results are shown in Fig.~\ref{fig:appendix_physical} and~\ref{fig:appendix_pollutants}, respectively. The second section presents the singular value decay plots.

\subsection{Singular-value decay}
\label{sec:appendix_sv}

\begin{figure}[h!]
    \centering
    \begin{subfigure}[b]{0.48\textwidth}
        \includegraphics[width=\textwidth]{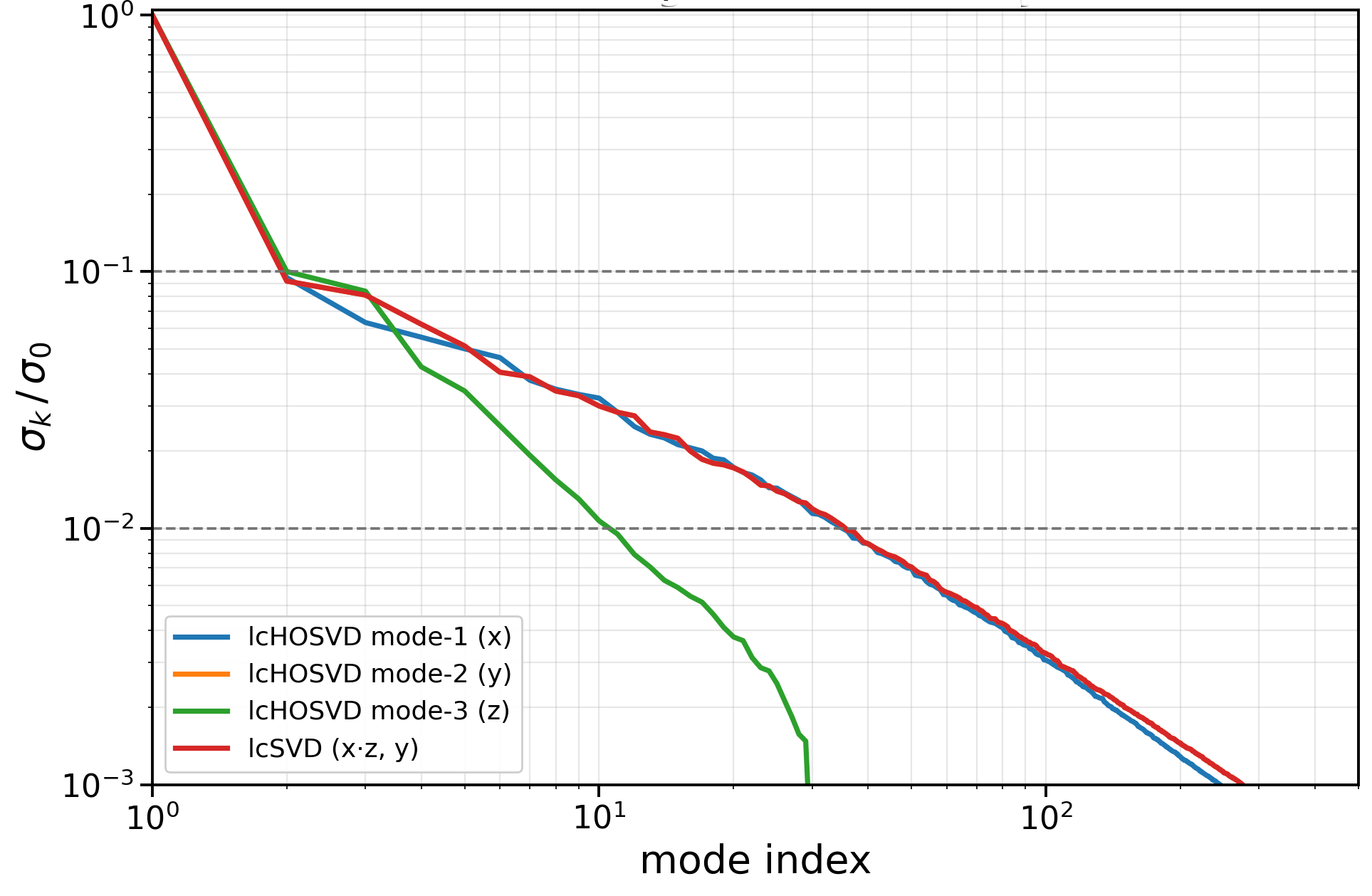}
        \caption{$u$}
    \end{subfigure}
    \hfill
    \begin{subfigure}[b]{0.48\textwidth}
        \includegraphics[width=\textwidth]{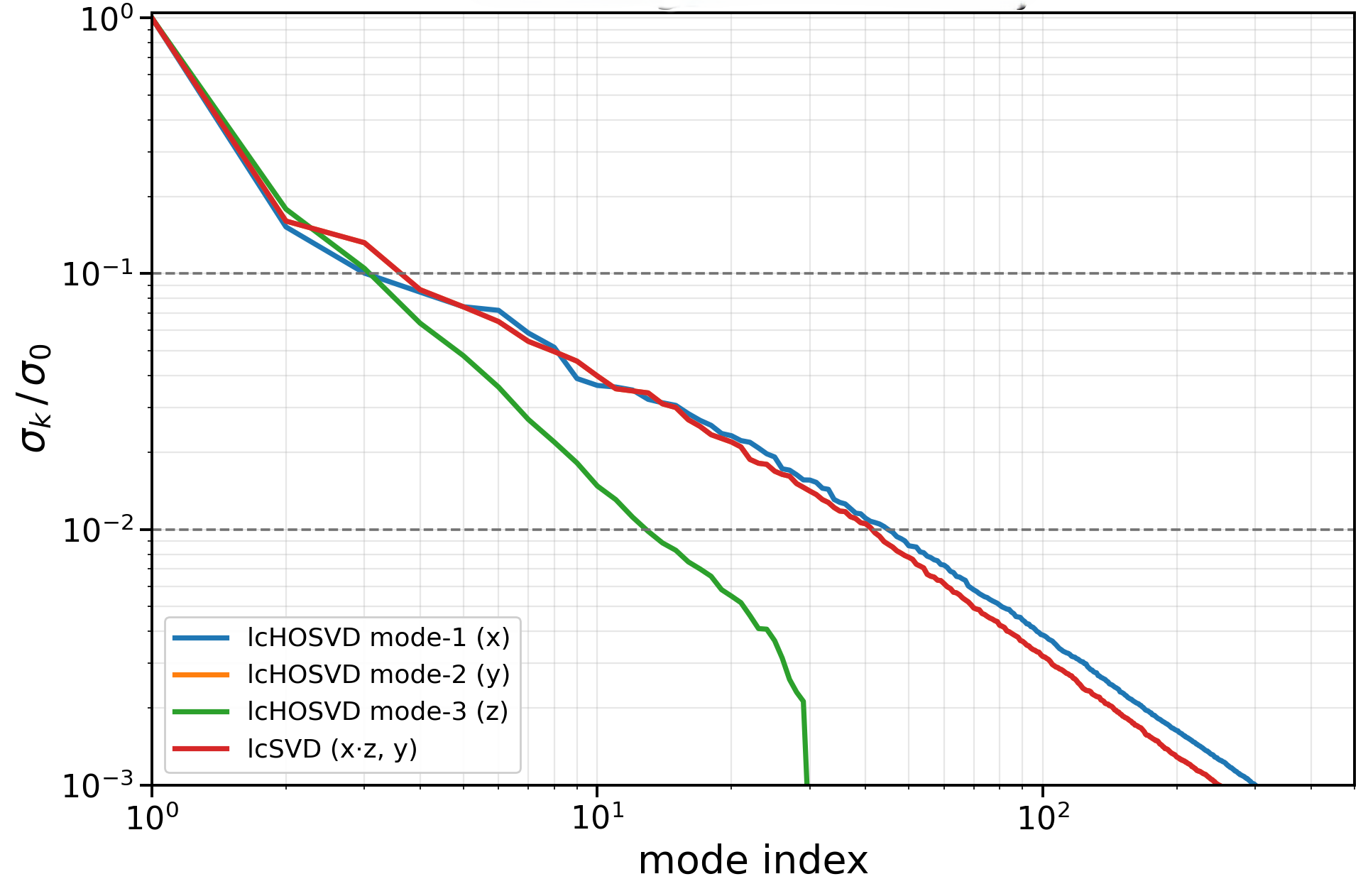}
        \caption{$v$}
    \end{subfigure}

    \vspace{0.5em}

    \begin{subfigure}[b]{0.48\textwidth}
        \includegraphics[width=\textwidth]{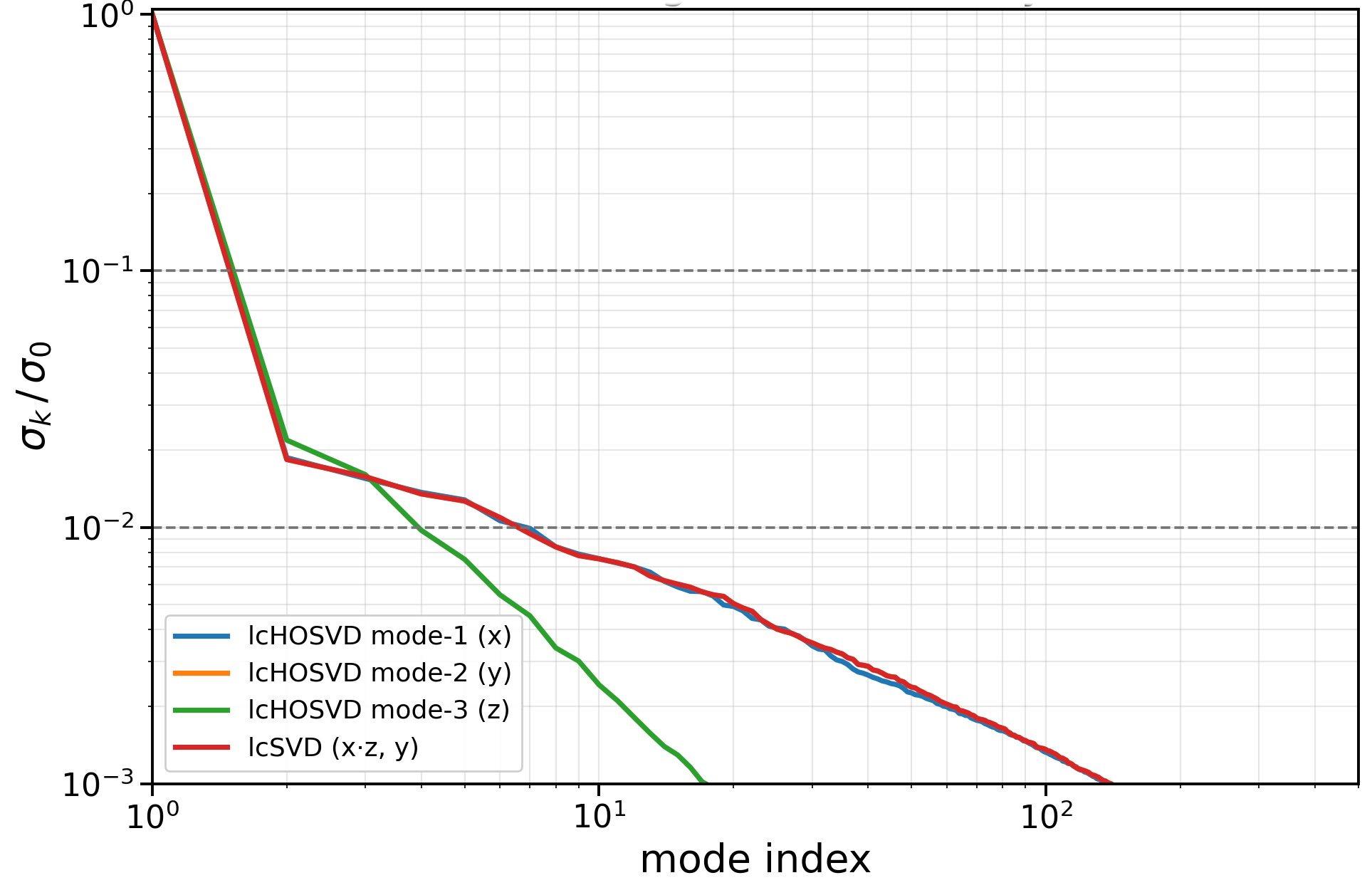}
        \caption{$w$}
    \end{subfigure}
    \hfill
    \begin{subfigure}[b]{0.48\textwidth}
        \includegraphics[width=\textwidth]{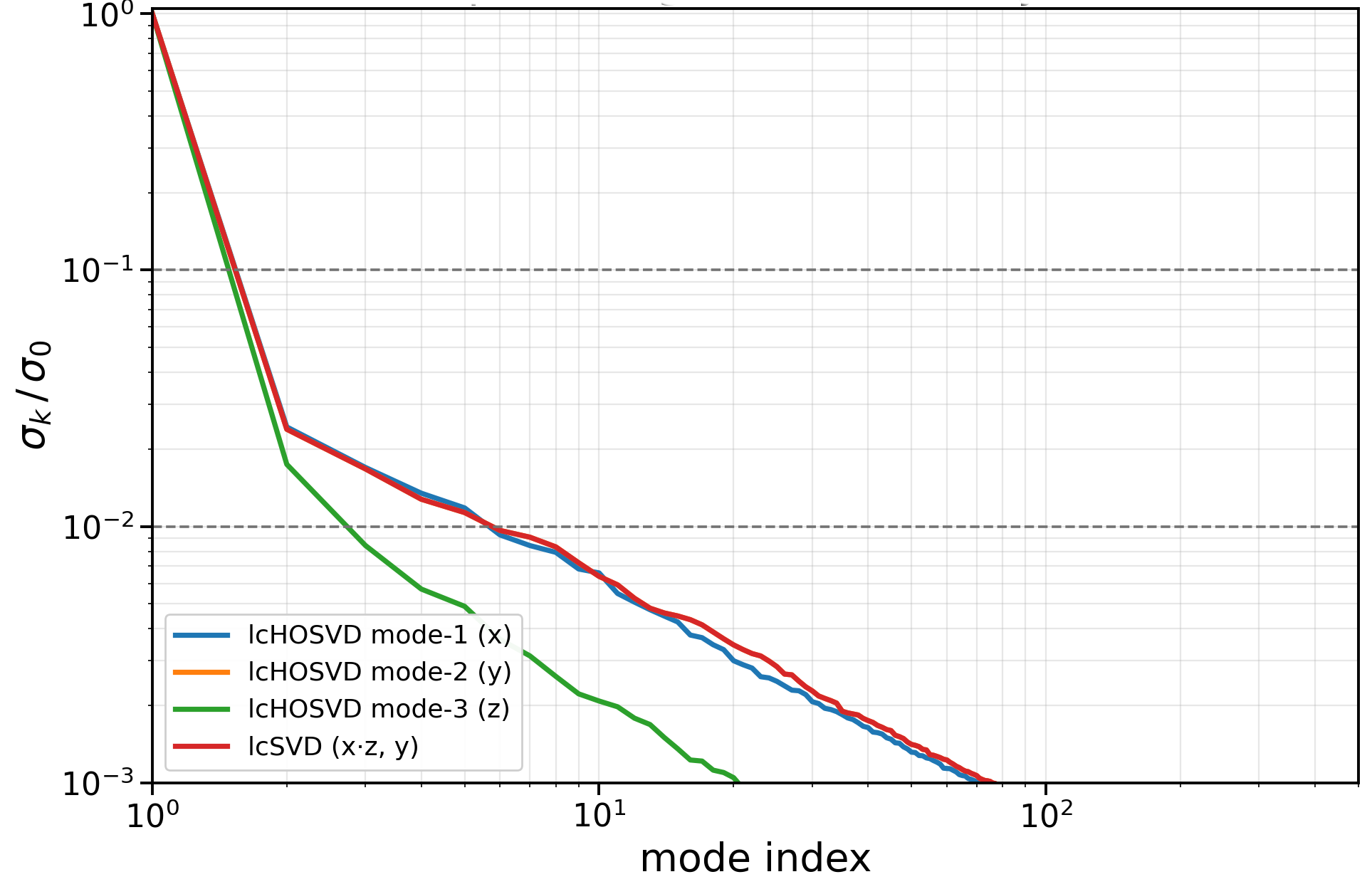}
        \caption{$p$}
    \end{subfigure}

    \vspace{0.5em}

    \begin{subfigure}[b]{0.48\textwidth}
        \includegraphics[width=\textwidth]{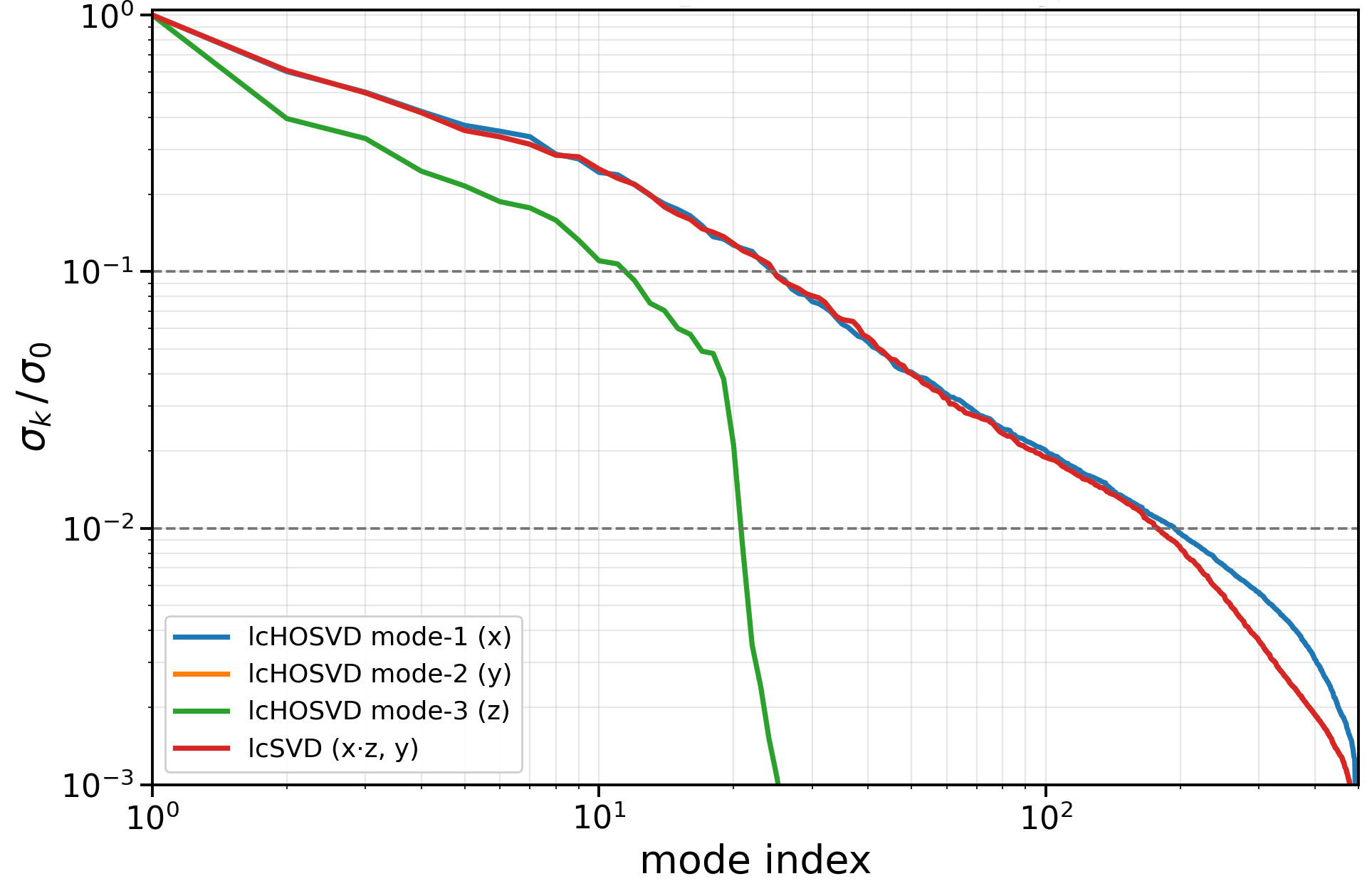}
        \caption{CO}
    \end{subfigure}
    \hfill
    \begin{subfigure}[b]{0.48\textwidth}
        \includegraphics[width=\textwidth]{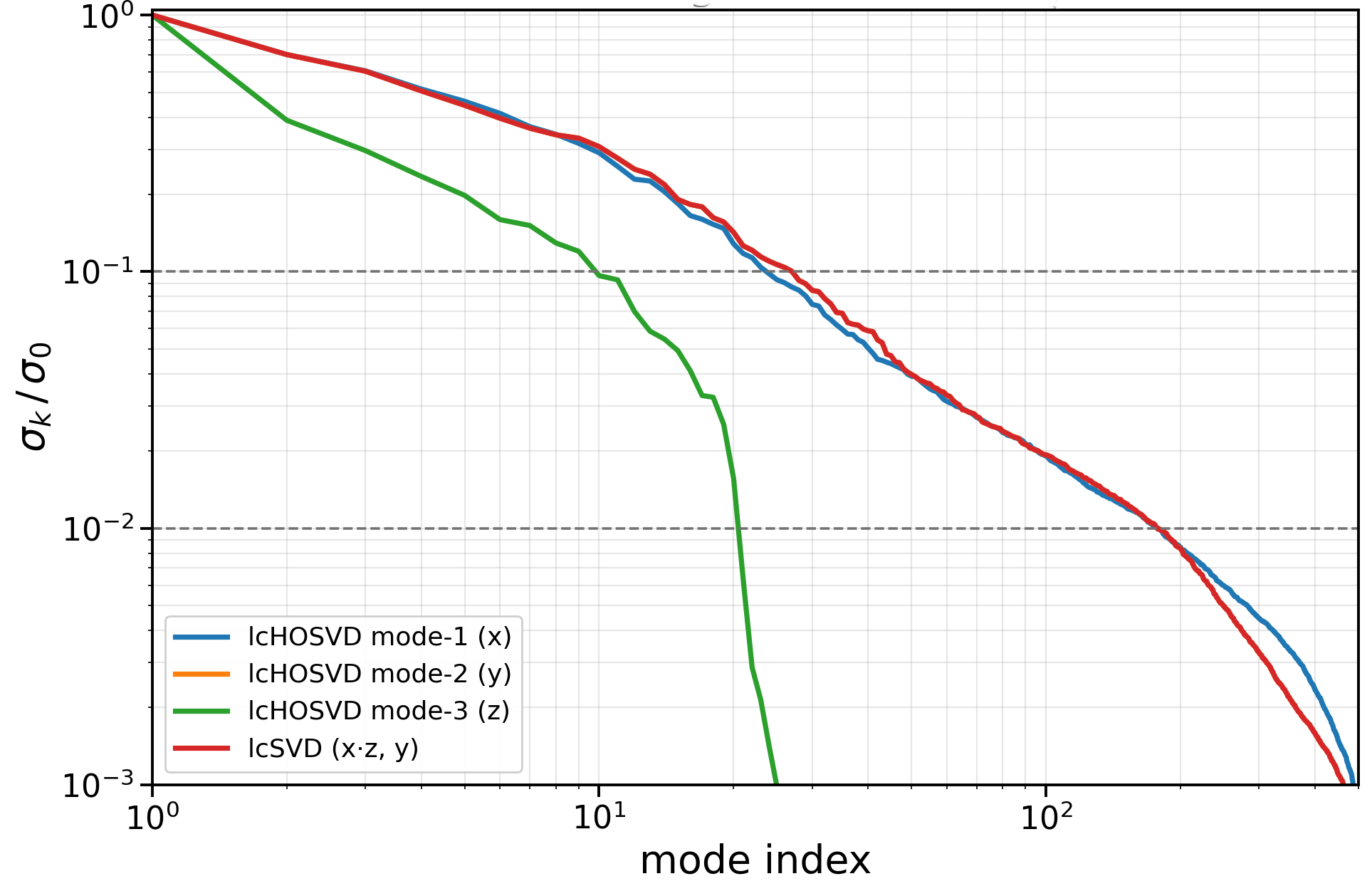}
        \caption{NO$_x$}
    \end{subfigure}

    \caption{Singular-value decay $\sigma_k/\sigma_0$ for the 
    remaining variables of the Vallecas dataset ($u$, $v$, $w$, 
    $p$, CO, NO$_x$).}
    \label{fig:sv_vallecas_remaining}
\end{figure}

\begin{figure}[h!]
    \centering
    \begin{subfigure}[b]{0.48\textwidth}
        \includegraphics[width=\textwidth]{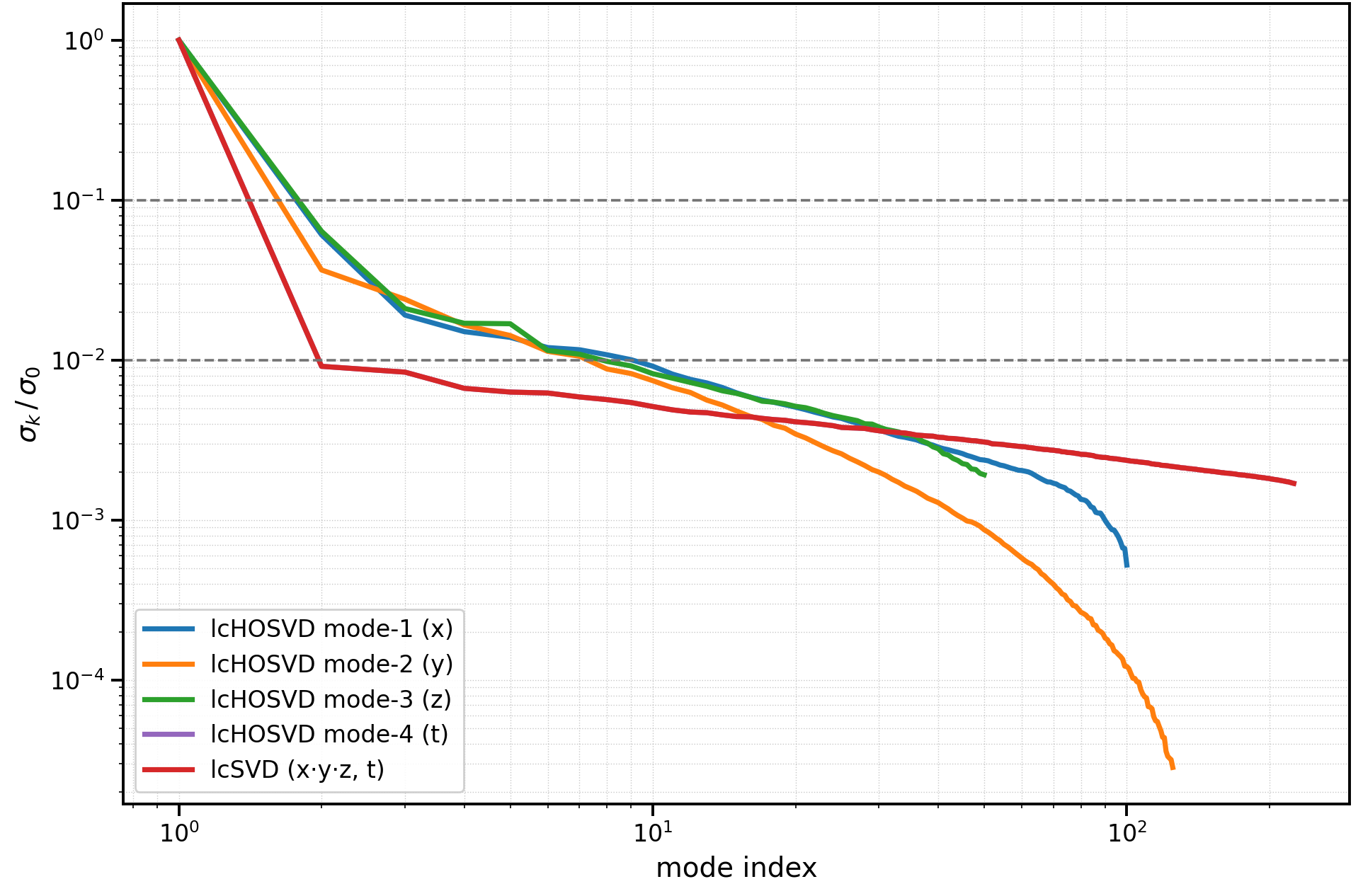}
        \caption{$u$}
    \end{subfigure}
    \hfill
    \begin{subfigure}[b]{0.48\textwidth}
        \includegraphics[width=\textwidth]{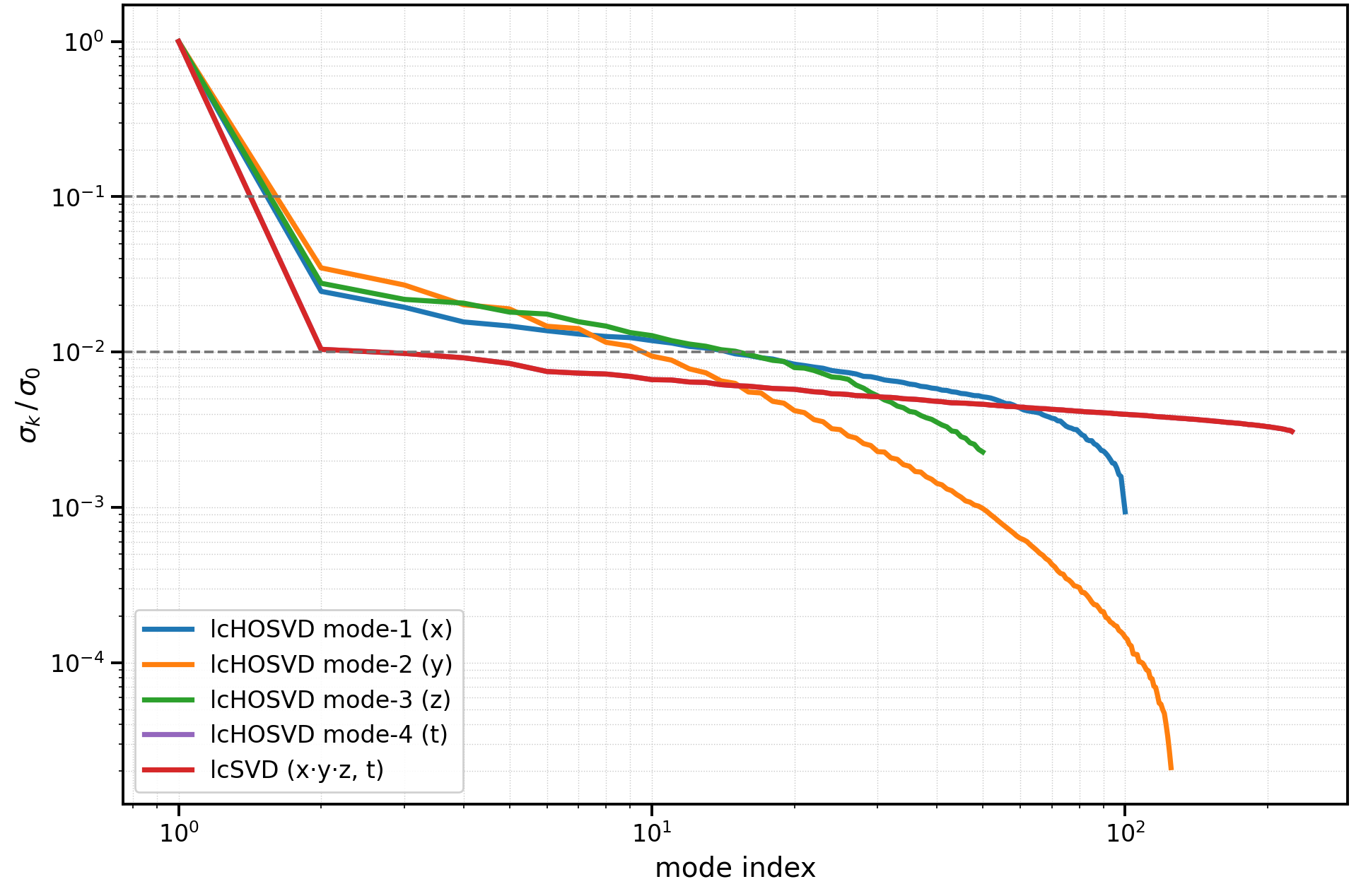}
        \caption{$v$}
    \end{subfigure}

    \vspace{0.5em}

    \begin{subfigure}[b]{0.48\textwidth}
        \includegraphics[width=\textwidth]{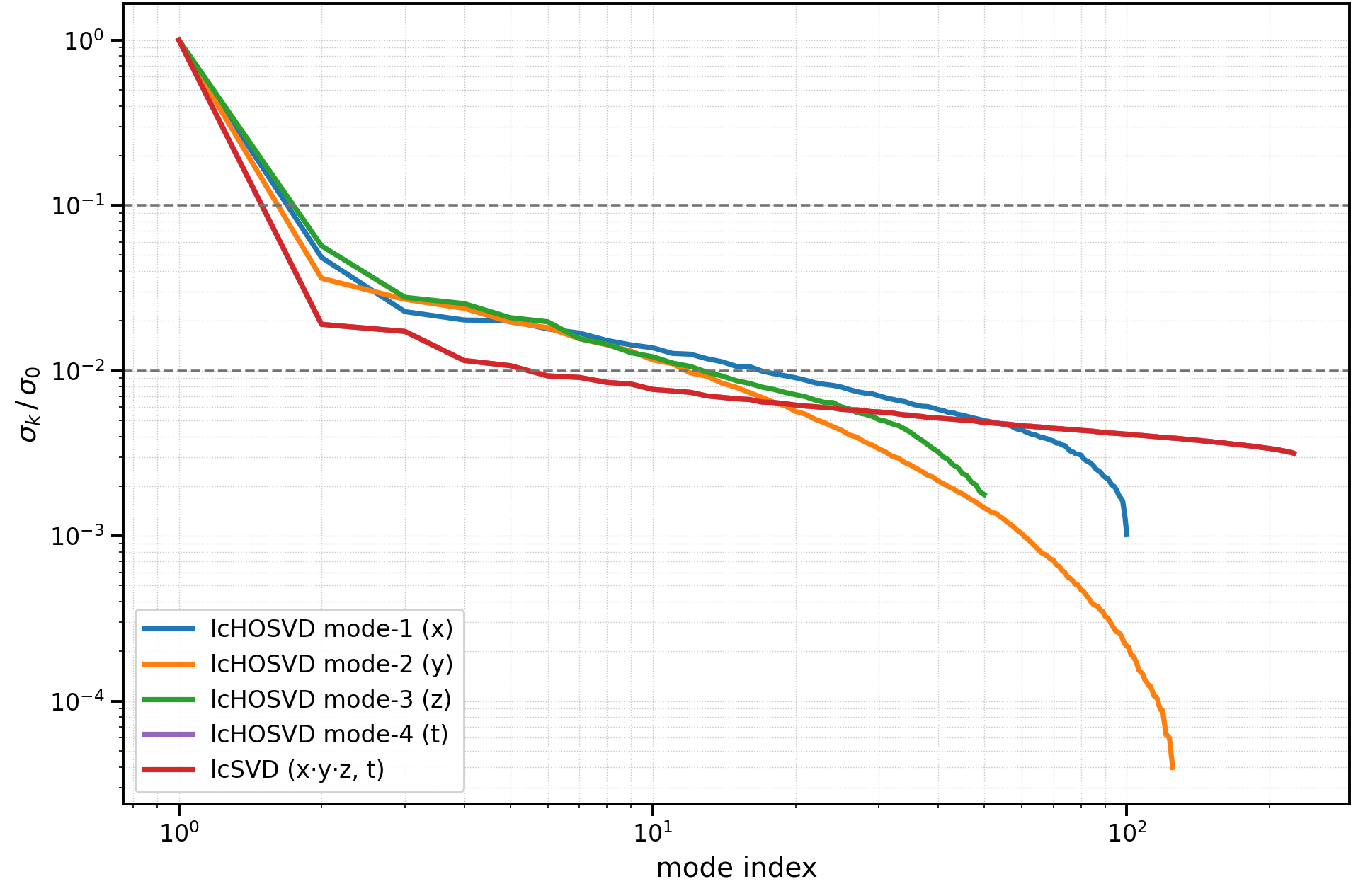}
        \caption{$w$}
    \end{subfigure}

    \caption{Singular-value decay $\sigma_k/\sigma_0$ for the two-building datset ($u$, $v$, $w$.)}
    \label{fig:sv_2bldg}
\end{figure}

The singular-value decay curves for all remaining variables are shown in Fig.~\ref{fig:sv_vallecas_remaining} 
and~\ref{fig:sv_2bldg}.

\end{appendices}
\newpage

\bibliography{jsbgf}
\end{document}